\documentclass[10pt,twocolumn,letterpaper]{article}

\usepackage[pagenumbers]{cvpr} 
\usepackage[accsupp]{axessibility}

\usepackage{graphicx}
\usepackage{amsmath}
\usepackage{color}
\usepackage{amssymb}
\usepackage{multirow}
\usepackage{booktabs}
\usepackage[super]{nth}

\usepackage[pagebackref,breaklinks,colorlinks]{hyperref}

\usepackage[capitalize]{cleveref}
\crefname{section}{Sec.}{Secs.}
\Crefname{section}{Section}{Sections}
\Crefname{table}{Table}{Tables}
\crefname{table}{Tab.}{Tabs.}

\def\cvprPaperID{2691} 
\def\confName{CVPR}
\def\confYear{2023}

\begin{document}

\title{Taylor3DNet: Fast 3D Shape Inference With Landmark Points Based Taylor Series}

\author{%
  Yuting Xiao\textsuperscript{\rm 1}\thanks{Equal contribution.} \quad
  Jiale Xu\textsuperscript{\rm 1}\footnotemark[1] \quad
  Shenghua Gao\textsuperscript{\rm 1,2,3}\thanks{Corresponding author.} \\
  \textsuperscript{\rm 1}ShanghaiTech University \\
  \textsuperscript{\rm 2}Shanghai Engineering Research Center of Intelligent Vision and Imaging \\
  \textsuperscript{\rm 3}Shanghai Engineering Research Center of Energy Efficient and Custom AI IC \\
  {\tt\small\{xiaoyt, xujl1, gaosh\}@shanghaitech.edu.cn} \\
}
\maketitle

\begin{abstract}
   Benefiting from the continuous representation ability, deep implicit functions can represent a shape at infinite resolution. However, extracting high-resolution iso-surface from an implicit function requires forward-propagating a network with a large number of parameters for numerous query points, thus preventing the generation speed. Inspired by the Taylor series, we propose Taylo3DNet to accelerate the inference of implicit shape representations. Taylor3DNet exploits a set of discrete landmark points and their corresponding Taylor series coefficients to represent the implicit field of a 3D shape, and the number of landmark points is independent of the resolution of the iso-surface extraction. Once the coefficients corresponding to the landmark points are predicted, the network evaluation for each query point can be simplified as a low-order Taylor series calculation with several nearest landmark points.  Based on this efficient representation, our Taylor3DNet achieves a significantly faster inference speed than classical network-based implicit functions. We evaluate our approach on reconstruction tasks with various input types, and the results demonstrate that our approach can improve the inference speed by a large margin without sacrificing the performance compared with state-of-the-art baselines.
\end{abstract}

\section{Introduction}
\label{sec:intro}
Deep neural-network-based implicit representation has gained much popularity in 3D shape modeling and reconstruction. Compared with explicit representations, \eg, voxels~\cite{girdhar2016learning,hane2017hierarchical,wu2016learning}, point clouds~\cite{qi2017pointnet,qi2017pointnet++,achlioptas2018learning,yin2019logan}, and meshes~\cite{wang2018pixel2mesh,kolotouros2019convolutional}, an implicit surface function represents a shape as a continuous level set in infinite resolution. In the inference phase, deep implicit functions~\cite{mescheder2019occupancy,peng2020convolutional,xu2019disn,saito2019pifu,park2019deepsdf,chabra2020deep} evaluate each query point to predict the occupancy or signed distance function (SDF) value and extract the iso-surface with a post-processing step such as Marching Cubes \cite{lorensen1987marching}.

However, the network evaluations for query points grow computationally unfriendly when extracting high-resolution surfaces required by real-world applications, since the forward propagation of the network is time-consuming and the amount of query points is extremely large. For example, extracting a mesh at $256$ resolution leads to $256^3$ query points in total. Previous work~\cite{mescheder2019occupancy,peng2020convolutional} proposes the Multiresolution IsoSurface Extraction (MISE) algorithm to sample query points hierarchically. Although it is effective in reducing the computational complexity, it needs to continuously subdivide the space from a lower resolution until reaching the target resolution, which cannot avoid producing a large number of query points when extracting surface at a high resolution.

\begin{figure*}[t]
\centering
\includegraphics[width=0.8\textwidth]{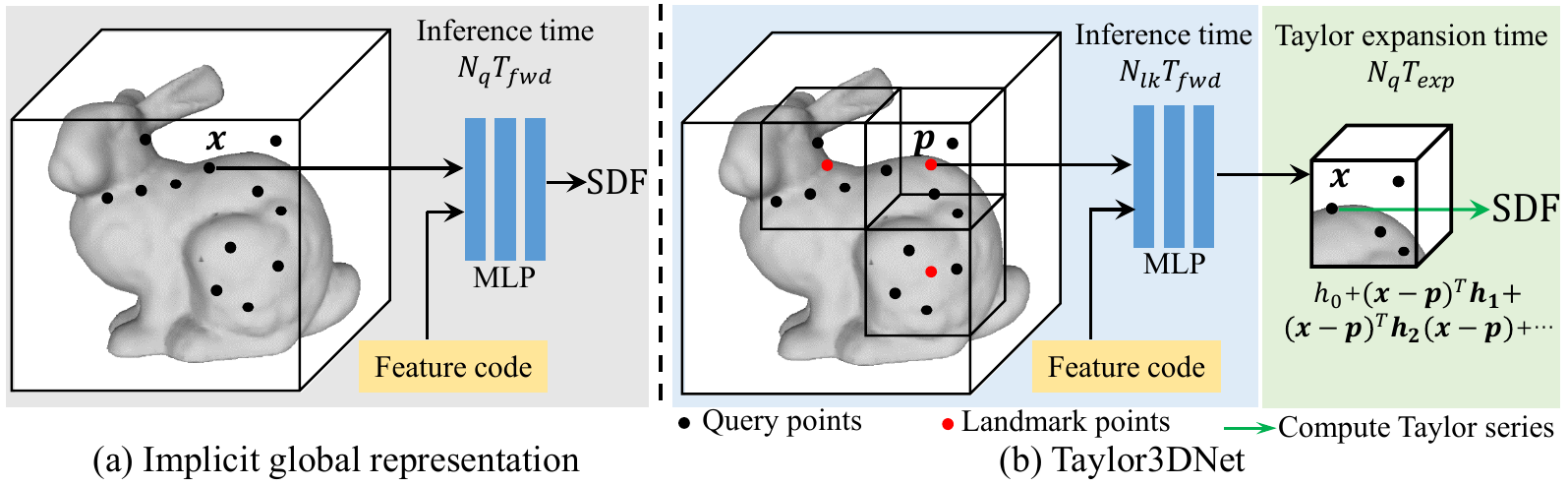}
\vspace{-2mm}
\caption{For high-resolution shape generation, the time consumed on feature encoding and marching cubes are significantly less than evaluating query points, so we focus on the point evaluation. Suppose that there are $N_q$ query points in the 3D space. (a) The classical deep implicit function forwards the network $N_q$ times totally. (b) Our method sample $N_{lk}$ $(N_{lk}\ll N_q)$ landmark points, each landmark point and its corresponding Taylor series represents a local region. The number of landmark point $N_{lk}$ is independent to the generation resolution, it does not change with the generation resolution. The time consumed on forwarding the MLP and computing Taylor series are $T_{fwd}$ and $T_{exp}$ $(T_{fwd}\gg T_{exp})$ respectively. So, the entire inference time of (a) and (b) are $N_qT_{fwd}$ and $N_{lk}T_{fwd}+N_qT_{exp}$ respectively and the former is much larger than the latter in the practical applications.}
\label{fig:teaser}
\vspace{-4mm}
\end{figure*}

Under the assumption that most of the local surface is smooth, we propose Taylor3DNet to approximate the SDF of a query point by leveraging a set of landmark points and the Taylor expansion technique. The landmark point indeed refers to the expansion point in the Taylor series. Each landmark point is associated with its Taylor series coefficients. With this representation, we can represent each shape with a set of landmark points and the corresponding Taylor series. Each landmark point and its Taylor series model a local implicit field. Our Taylor3DNet is trained to predict the Taylor series coefficients of the landmark points, thus the evaluation for each query point can be simplified as calculating the Taylor series with several nearest landmark points.

As \cref{fig:teaser} shows, classical implicit functions evaluate each query point by propagating an MLP network, and $N_q$ query points require $N_q$ propagations. As the resolution for surface extraction increases, the number of query points grows cubically. In comparison, our method only needs to evaluate $N_{lk}$ landmark points to predict their coefficients of the Taylor series and ensure that the union of the local regions represented by the landmark points can cover the overall shape, which indicates that $N_{lk}$ is independent of the generation resolution. Besides, we propose a coarse-to-fine landmark points sampling strategy in the inference phase to ensure most of the sampled landmark points are located near the surface so as to preserve the shape details and minimize the number of landmark points.

Based on the fact that $N_{lk}\ll N_q$ and $N_{lk}$ is independent of the generation resolution, the network propagation time for landmark points is constant. And computing the Taylor series for query points is very efficient. As a result, we achieve a significant acceleration in the shape inference speed with our representation. Furthermore, it also maintains the representation capacity of deep neural-network-based implicit functions.

We test our shape representation on toy data to demonstrate what it actually learns. Moreover, we evaluate our Taylor3DNet on the ShapeNet~\cite{chang2015shapenet} dataset to show its performance on shape reconstruction tasks with various input data types. Our contributions can be summarized as:

\begin{itemize}
    \item We propose Taylor3DNet to represent 3D shapes based on the landmark points and low-order Taylor series, which is friendly to fast shape inference.
    \item We propose a coarse-to-fine landmark points sampling strategy in the inference phase, which is independent of the resolution of mesh generation.
    \item We evaluate on reconstruction tasks with various input types and demonstrate that it can significantly accelerate the inference speed while maintaining the performance compared with state-of-the-art baselines.
\end{itemize}

\section{Related Work}
\noindent\textbf{3D Shape Reconstruction.} 
3D shape reconstruction from various input types has been extensively studied. Given a single-view image, learning-based methods~\cite{choy20163d,fan2017point,groueix2018papier,wang2018pixel2mesh,xie2019pix2vox} predict an explicit voxel grid, point cloud or mesh. Compared with explicit representations, deep implicit functions have advantages in memory efficiency and representation capacity, thus are preferred in reconstruction tasks~\cite{mescheder2019occupancy,saito2019pifu,xu2019disn,thai20213d}. Unlike the ill-posed single-view reconstruction, reconstruction from point clouds or coarse voxel grids leads to finer shapes thanks to their inherent 3D priors. Given an input point cloud, traditional optimization-based methods, \eg, Moving Least Square \cite{alexa2003computing} and Poisson Surface Reconstruction~\cite{kazhdan2006poisson,kazhdan2013screened}, and deep optimization-based methods such as SAL~\cite{Atzmon_2020_CVPR}, IGR~\cite{gropp2020implicit} and Neural Splines~\cite{williams2021neural} can fit a compact surface with fine details, but the optimization process for each shape often takes a long time. Learning-based implicit methods~\cite{mescheder2019occupancy,peng2020convolutional,chibane2020implicit,Tang_2021_ICCV,Venkatesh_2021_ICCV} need ground truth SDF or occupancy values for supervision, but they can inference in a feed-forward manner. Despite the representation capacity, implicit functions require a post-processing step, \eg, Marching Cubes, for iso-surface extraction, which is often time-consuming at high resolutions.

\begin{figure*}[t]
  \centering
  \includegraphics[width=0.9\textwidth]{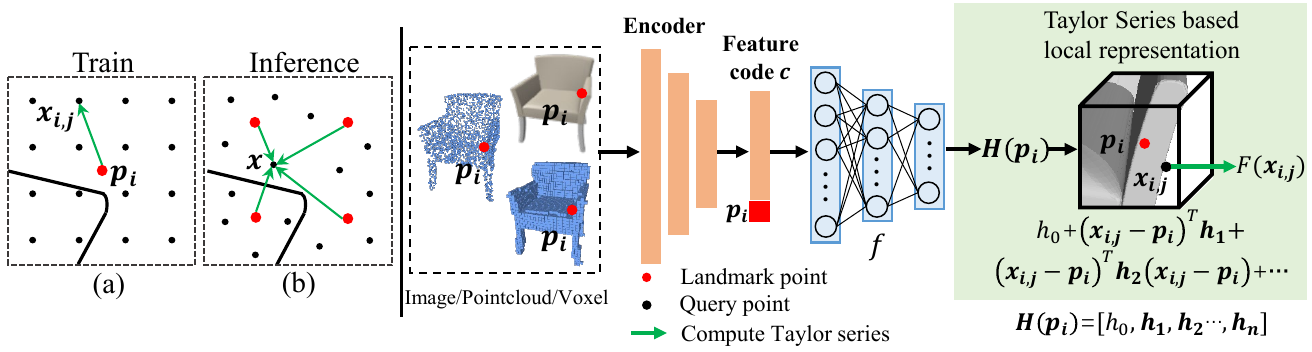}
  \vspace{-3mm}
  \caption{Overview of our Taylor3DNet. The input data can be single-view images, point clouds, or voxel grids. In the training stage, we sample a set of query points $\{\boldsymbol{x}_{i,j}\}_{j=1}^N$ near each landmark point $\boldsymbol{p}_i$ and train the network $f$ to predict the Taylor series coefficients $\boldsymbol{H}(\boldsymbol{p}_i)$ of $\boldsymbol{p}_i$ by computing the signed distances of the query points. In the inference stage, we sample a set of landmark points and predict their Taylor series coefficients using $f$, and compute the signed distance of arbitrary query point $\boldsymbol{x}$ by the weighted average of the Taylor series corresponding to multiple nearest landmark points.}
  \label{fig:exp_examples}
  \vspace{-5mm}
\end{figure*}

\noindent\textbf{Global Shape Representations.} 
Global shape representations represent a shape with a single implicit function. They often exploit an auto-encoder (AE) architecture~\cite{mescheder2019occupancy,chen2019learning}, in which an encoder maps the input shape observation into a latent code, and a decoder recovers the shape from the latent code. DeepSDF \cite{park2019deepsdf} adopts an auto-decoder architecture instead, which randomly samples a latent code from a Gaussian distribution for each shape and optimizes it with gradient descent during training. A series of work~\cite{duan2020curriculum,mu2021sdf,zheng2021deep,deng2021deformed,shan2021ellipsdf} has improved this framework and shows impressive shape modeling results. Global shape representations have good shape completion ability, and the learned latent space is convenient for shape interpolation and generation. However, they struggle to recover fine details. Thus, some recent work~\cite{sitzmann2020implicit,tancik2020fourier} proposes to use periodic activation functions for better preserving high-frequency details.

\noindent\textbf{Local Shape Representations.}
local shape representations decompose a holistic shape into local parts to better model shape details. Some methods learn to decompose the shape into local parts automatically, and represent the parts with primitives~\cite{Deng_2020_CVPR,ren2021csg,Paschalidou_2021_CVPR,Yao_2021_ICCV}, quartics~\cite{Paschalidou_2019_CVPR,Yavartanoo_2021_ICCV} or 3D Gaussians~\cite{genova2020local,genova2019learning}. 
Another routine seeks to divide the 3D space into local patches. DeepLS \cite{chabra2020deep} uniformly divides the 3D space into voxel grids and assigns a latent code as well as a DeepSDF decoder to each voxel. LGCL \cite{yao20213d} adopts a similar idea except that the space partition is defined by a set of key points. LIG \cite{jiang2020local} train a part auto-encoder to learn an embedding of local crops of 3D shapes.

\section{Method} 

In this section, we introduce the formulation of our implicit 3D shape representation with landmark points based Taylor series, the training scheme and the coarse-to-fine landmark points sampling strategy in the inference phase.

Given partial 3D observations as input, \eg, images, point clouds, or voxel grids, classical implicit-function-based reconstruction methods first learn a latent code $\boldsymbol{c}$, and then use $\boldsymbol{c}$ and a neural network $f$ to predict the SDF or occupancy value of a query point $\boldsymbol{x} \in \mathbb{R}^3$, producing an implicit field $F(\boldsymbol{x}) = f(\boldsymbol{x}, \boldsymbol{c})$.

Deep implicit functions represent the shape surface as a continuous level set $\{\boldsymbol{x}|F(\boldsymbol{x})=l\}$, where $l$ is a scalar representing the decision boundary. Despite the continuous shape representation capacity, extracting the iso-surface with this representation requires feeding a set of query points into the network $f$ to locate the surface with the network outputs. Denoting the time of forwarding the network $f$ once as $T_{fwd}$ and the number of query points as $N_q$, then the time cost for evaluating all query points is $N_qT_{fwd}$. $T_{fwd}$ is indeed a nonnegligible period of time due to the large amount of parameters of the implicit function. When extracting high-resolution surfaces, $N_q$ grows very large and leads to a high total time cost $N_qT_{fwd}$, thus preventing the generation speed significantly.


\subsection{Taylor-based Implicit Function Formulation}

An overview of our Taylor3DNet is shown in \cref{fig:exp_examples}. Given the signed distance field $F(\boldsymbol{x})$ of a shape, we aim to utilize a set of landmark points $\{\boldsymbol{p}_i\}_{i=1}^{N_{lk}}$ and associated  Taylor series to approximate $F(\boldsymbol{x})$, where $N_{lk}$ is the number of landmark points. For each landmark point $\boldsymbol{p} \in \mathbb{R}^3$, we leverage the Taylor series at point $\boldsymbol{p}$ to model a local region of $F(\boldsymbol{x})$ near $\boldsymbol{p}$. The local implicit function of the region near $\boldsymbol{p}$ is formulated as:
\begin{equation}
F(\boldsymbol{x}; \boldsymbol{p}) = h_0 + (\boldsymbol{x} - \boldsymbol{p})^T\boldsymbol{h_1} + \frac{1}{2!}(\boldsymbol{x} - \boldsymbol{p})^T\boldsymbol{h_2}(\boldsymbol{x} - \boldsymbol{p}) + \cdots
\label{eq:taylor}
\end{equation}
where $h_0 = F(\boldsymbol{p})\in \mathbb{R}$ is the 0-order term, \ie, the signed distance of point $\boldsymbol{p}$, $\boldsymbol{h}_1 = \nabla F(\boldsymbol{x}) \in \mathbb{R}^3$ is the gradient of $F(\boldsymbol{x})$ derived at $\boldsymbol{x}=\boldsymbol{p}$ and the $\boldsymbol{h}_2 \in \mathbb{R}^{3\times 3}$ is the Hessian matrix of $F(\boldsymbol{x})$ derived at $\boldsymbol{x}=\boldsymbol{p}$. We will show that the 2-order Taylor series is powerful enough for our shape representation in \cref{sec:capacity}.

$F(\boldsymbol{x}; \boldsymbol{p})$ is a Taylor series and is irrelevant to neural networks. As shown in \cref{fig:exp_examples}, the coefficients $\boldsymbol{H}(\boldsymbol{p}) = [h_0, \boldsymbol{h}_1, \boldsymbol{h}_2\cdots ,\boldsymbol{h}_n]$ of Taylor series $F(\boldsymbol{x}; \boldsymbol{p})$ are predicted by a network $f$:
\begin{equation}
    \boldsymbol{H}(\boldsymbol{p}) = f(\boldsymbol{p}, \boldsymbol{c})
    \label{equ:3}
\end{equation}

With this representation, we only need to feed the $N_{lk}$ landmark points into the network $f$ to predict their coefficients of the local Taylor series. Then we can discard the network and use the landmark points and their corresponding Taylor series to represent the whole shape. For any query point $\boldsymbol{x}$, we can infer its SDF value by computing the Taylor series in \cref{eq:taylor} with a landmark point $\boldsymbol{p}$ nearby. 

When it comes to high-resolution shape reconstruction, the number of landmark points $N_{lk}$ is significantly less than the number of query points $N_q$ $(N_q\gg N_{lk})$, since the landmark points are sampled independently while the number of query points grows quickly as the generation resolution increases. Denoting the time of computing the Taylor series once as $T_{exp}$, the total points evaluation time cost of our method is $N_{lk}T_{fwd}+N_qT_{exp}$. Based on the facts that $N_{lk}\ll N_{q}$ and $T_{fwd}\gg T_{exp}$, we can infer that $N_qT_{fwd} \gg N_{lk}T_{fwd}+N_qT_{exp}$, which indicates the acceleration effect of our representation. Besides, our method reduces the inference time by a factor of $(N_{lk}T_{fwd}+N_q T_{exp})/\  N_q T_{fwd}=N_{lk}/\ N_q+ T_{exp} /\ T_{fwd}$. When generating shapes with higher resolutions ($N_{lk}/N_{q}\downarrow$) or using neural networks with more parameters ($T_{exp}/T_{fwd}\downarrow$), our method will lead to a more significant speedup.


\subsection{Training Phase}
In the training phase, our Taylor3DNet learns to predict the coefficients of the Taylor series for each landmark point with a network $f$, and $f$ can be implemented with the architecture of any classical implicit function-based model. 

For each landmark point $\boldsymbol{p}_i\in\mathbb{R}^3$, we concatenate it with the feature code $\boldsymbol{c}\in\mathbb{R}^d$ obtained from the input data and feed them into the network $f$, and $f$ outputs the coefficients $\boldsymbol{H}(\boldsymbol{p}_i)$ as shown in \cref{equ:3}. To supervise the network $f$, we randomly sample $M$ landmark points $\{\boldsymbol{p}_i\}_{i=1}^M$ in the bounding volume of the target object, and a grid of query points $\{\boldsymbol{x}_{i,j}\}_{j=1}^N$ around each landmark point $\boldsymbol{p}_i$, where $\boldsymbol{x}_{i,j} \in \mathbb{R}^3$ is the $j$-th query point corresponding to the $i$-th landmark point $\boldsymbol{p}_i$. We then use $\boldsymbol{H}(\boldsymbol{p}_i)$ to compute the signed distances of these query points and supervise them with ground truth SDF values.

Directly applying the SDF regression loss is not suitable for the shape reconstruction task, because the SDF errors at the points far from the surface are not significant for the reconstruction results. In order to make the network focus on the reconstruction of the regions near the surface, we apply a sigmoid transformation on the signed distance field: $\sigma(s) = 1/(1+e^{-\alpha s})$, where $s$ is a signed distance and $\alpha$ is a scaling hyperparameter. After transformation, the SDF values of the points far from the surface would be smoothed and the diversity reduction would make the prediction easier. Besides, applying the TSDF (Truncated Signed Distance Field) can achieve similar effects but it suffers from the existence of the non-differentiable points, which is hard to be approximated by lower-order Taylor series.

Our Taylor3DNet is trained by minimizing the following loss function:
\begin{equation}
\mathcal{L} = \frac{1}{MN}\sum_{i=0}^M\sum_{j=0}^N \mathcal{L}_{ce}(\sigma(\hat{F}(\boldsymbol{x}_{i,j}; \boldsymbol{p}_i)), \sigma(F(\boldsymbol{x}_{i,j})))
\label{eq_loss}
\end{equation}
where $\mathcal{L}_{ce}$ denotes the cross-entropy loss, $\hat{F}(\boldsymbol{x}_{i,j}; \boldsymbol{p}_i)$ is the predicted signed distance of query point $\boldsymbol{x}_{i,j}$ with respect to the landmark point $\boldsymbol{p}_i$ and $F(\boldsymbol{x}_{i,j})$ is the ground truth signed distance of $\boldsymbol{x}_{i,j}$.

\subsection{Inference Phase}
In the inference phase, we propose a coarse-to-fine strategy to sample landmark points and extract the iso-surface efficiently.

Firstly, we sample a low-resolution grid of landmark points in the 3D volume to localize the coarse area of the target surface. For each landmark point, we compute its signed distance with the Taylor series corresponding to it. Then we use an inside distance threshold $\epsilon_{in}$ and an outside threshold $\epsilon_{out}$ to classify the landmark points into three classes: (i) outside landmark points far from the surface, (ii) inside landmark points far from the surface, and (iii) landmark points near the surface. we aim to sample denser landmark points only in the regions close to the surface, which can be denoted as:
\begin{equation}
    \mathcal{R}=\{R(\boldsymbol{p}_i)|\epsilon_{in}\leq \sigma(\hat{F}(\boldsymbol{p}_i, \boldsymbol{p}_i))\leq \epsilon_{out}\}
\end{equation}
where $R(\boldsymbol{p}_i)$ is the local cube region corresponding to the landmark point $\boldsymbol{p}_i$, and $\hat{F}(\boldsymbol{p}_i, \boldsymbol{p}_i)$ is the signed distance of $\boldsymbol{p}_i$, which is exactly the 0-order term of the Taylor series at $\boldsymbol{p}_i$. For the query points not in $\mathcal{R}$, we can directly set their SDF the same as its nearest coarse landmark points.

\begin{figure}[t]
  \centering
  \includegraphics[width=\linewidth]{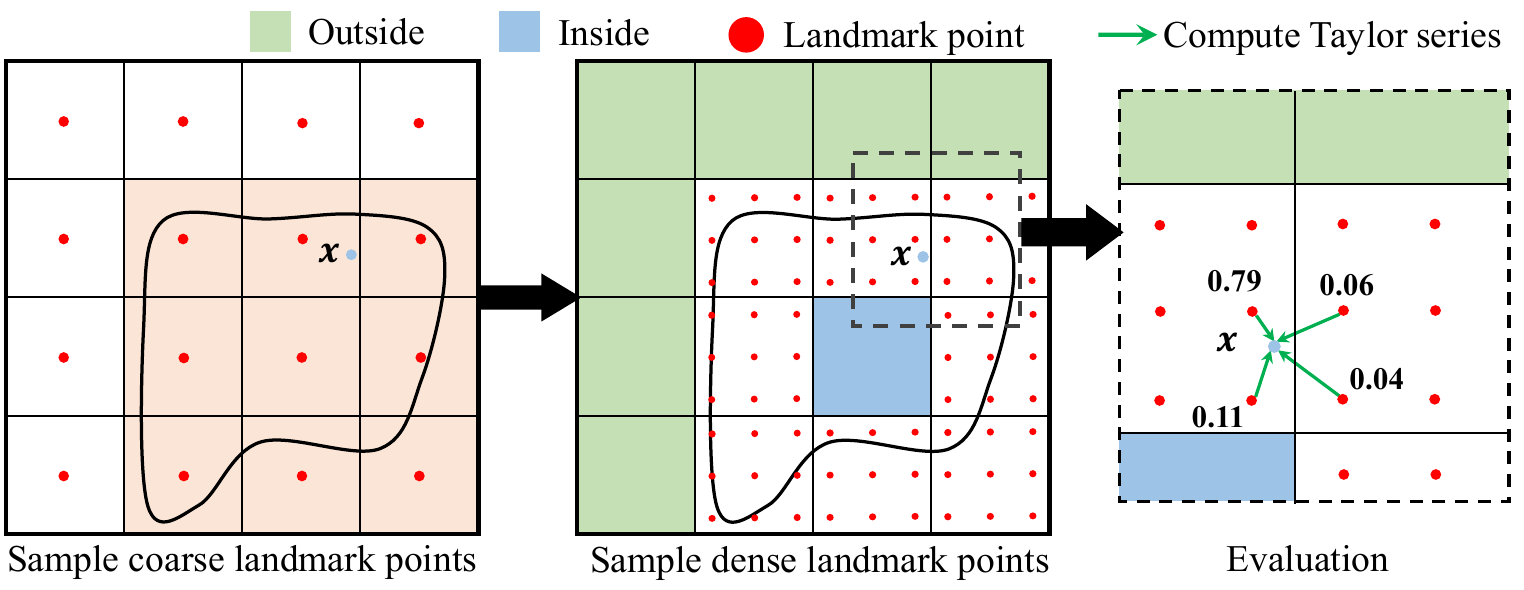}
  \vspace{-6mm}
  \caption{An example to introduce the coarse-to-fine strategy for landmark point sampling in the inference phase. The signed distance of a query point is estimated by the 4 nearest landmark points with the weights are $0.79$, $0.11$, $0.06$, and $0.04$.}
  \label{fig:test_phase}
  \vspace{-5mm}
\end{figure}

Then, we subdivide each local cube region $R(\boldsymbol{p}_i)$ as a $2^3$ grid and sample finer landmark points in this grid. We then use the set of fine landmark points to evaluate the query point in $\mathcal{R}$. With this strategy, we can infer the signed distance of arbitrary query points in space, as shown in \cref{fig:test_phase}.

However, only applying single nearest landmark point produces discontinuous reconstruction results at the middle region of two landmark points. To alleviate this weakness, we adopt a weighted average operation among the Taylor series of $k$ nearest landmark points to compute $F(\boldsymbol{x})$ for smoothing. Denoting $\mathcal{P}$ as a set of closest landmark points of query point $\boldsymbol{x}$, the signed distance $F(\boldsymbol{x})$ is computed by:
\begin{equation}
F(\boldsymbol{x}) = {\sum}_{\boldsymbol{p}_i\in \mathcal{P}}\omega(\|\boldsymbol{x}-\boldsymbol{p}_i\|, \theta)F(\boldsymbol{x}; \boldsymbol{p}_i)
\label{eq:avg}
\end{equation}
where $\omega$ denotes the the Softmin function: $\omega(d_i, \theta)= \text{exp}(-\theta d_i)/\sum_j \text{exp}(-\theta d_j)$, $d_i$ denotes the Euclidean distance between the query point $\boldsymbol{x}$ and the landmark point $\boldsymbol{p}_i$.

\begin{figure*}[t]
  \centering
  \includegraphics[width=0.85\textwidth]{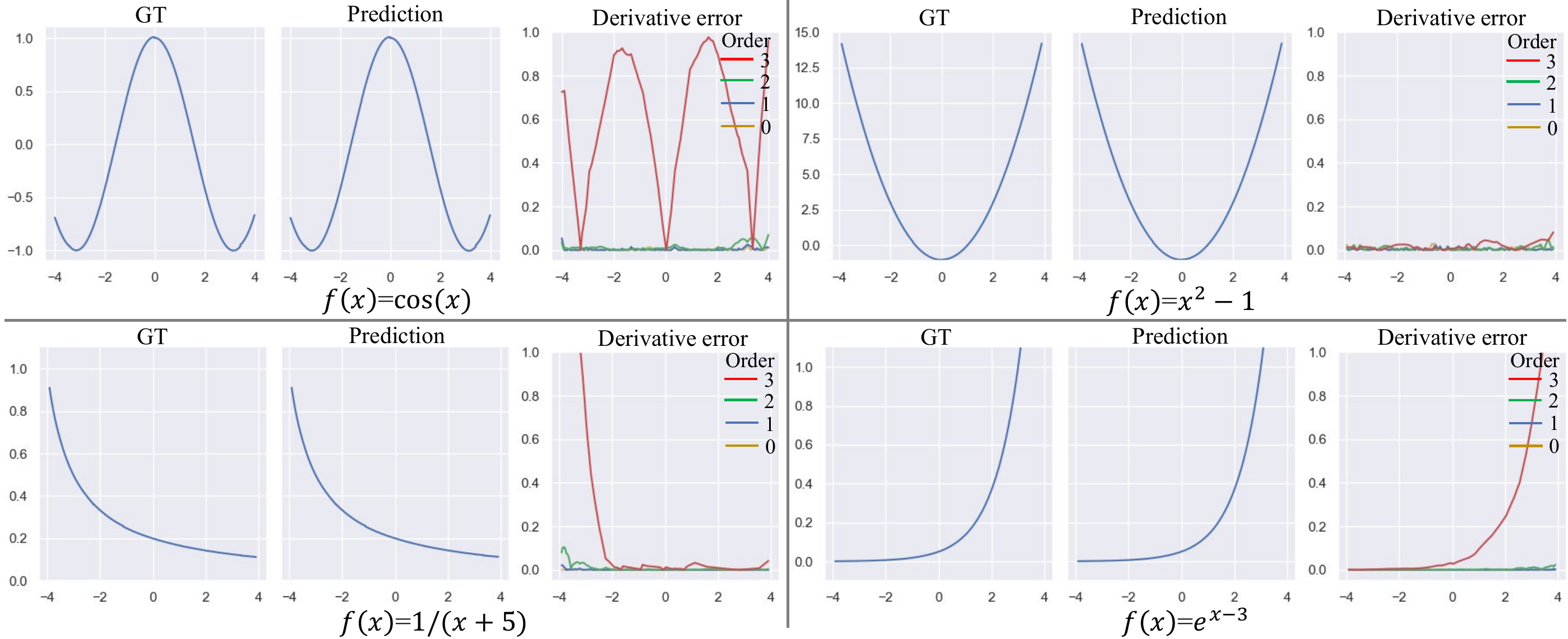}
  \vspace{-3mm}
  \caption{The experiments on the toy data show that our method can well approximate the target function.}
  \label{fig:toydata}
  \vspace{-4mm}
\end{figure*}

\section{Experiments}
In this section, we first conduct experiments on the toy data to demonstrate the representation capacity of the Taylor series. Then we demonstrate that Taylor3DNet speed up the inference while preserving the performance. Finally, we conduct ablation studies to discuss some crucial issues.

\noindent\textbf{Dataset.}
We conduct experiments on 13 shape categories of the ShapeNet-v1~\cite{chang2015shapenet} dataset. We follow the preprocessing pipeline of OccNet~\cite{mescheder2019occupancy} to obtain watertight meshes and normalize them into a unit cube. We adopt the same train/val/test split as~\cite{mescheder2019occupancy} for a fair comparison. 

\noindent\textbf{Metrics.} 
Following OccNet~\cite{mescheder2019occupancy} and ConvOccNet\cite{peng2020convolutional}, we adopt three metrics: (i) volumetric IoU (higher is better) which is computed by uniformly sample 100K points in the bounding volume of each mesh, (ii) Chamfer-$L_1$ distance (lower is better) which is computed by randomly sampling 100k points on the mesh surface and scaled by 10, and (iii) F-Score (higher is better) which is calculated with the threshold $\tau=0.01$.

\noindent\textbf{Training Details.}
We randomly sample $m_1$ and $m_2$ landmark points ($k_1:k_2=1:3$) in the mesh bounding volume and near the surface, respectively. Then we evenly sample a $5^3$ grid of query points in a $0.08^3$ cube centered at each landmark point. We use the Adam optimizer with $\beta_1 = 0.9$ and $\beta_2 = 0.999$. The learning rate is initialized as $1\times 10^{-3}$ and divided by 10 for two times during training. The training takes about two days on a single Titan V GPU.

\noindent\textbf{Inference Details.} 
In the inference phase, our coarse-to-fine strategy firstly samples a $16^3$ grid of coarse landmark points and uses the thresholds $\epsilon_{out}=0.98$ and $\epsilon_{in}=0.02$ to localize the regions near the surface. Then, we subdivide each cube region near the surface into a $2^3$ grid to sample fine landmark points. And we evaluate each query point with the four nearest landmark points.


\subsection{Experiments on Toy Data}
In this section, we utilize four 1D simple functions to analyze whether the neural network can produce the coefficients of their Taylor series: (i) $f(x)=\cos(x)$, (ii) $f(x)=x^2-1$, (iii) $f(x)=1/(x+5)$, and (iv) $f(x)=e^{x-3}$. The cosine function's Taylor series has an infinite number of higher-order terms while the others' have a finite number of higher-order terms.

We use a network to predict the coefficients of the Taylor series, \ie, the different orders of derivatives, at a given point. To train the network, we randomly sample 128 landmark points within the range $[-4.0, 4.0]$. For each landmark point $p_i$, we uniformly sample 32 query points in the range $[p_i-0.4, p_i+0.4]$. We feed the landmark point $p_i$ into the network to infer the different orders of derivatives at $p_i$ and use the Taylor series to predict the function values at the sampled query points. The network is then trained by minimizing the $L_1$ loss.

In the inference phase, we sample 8 landmark points within the range $[-4.0, 4.0]$ and try to recover the original function with them. We also use the network to predict the derivatives of 1000 evenly sampled points to evaluate the inference error of derivatives. The results are shown in \cref{fig:toydata}. 

We can observe that the network can reconstruct the original function precisely using the landmark points and their coefficients of the Taylor series. Besides, the lower-order derivative terms inferred by the network are significantly accurate, while the prediction of higher-order terms is difficult and shows large errors. The reason is that the higher-order terms of the Taylor series contribute little to the reconstruction results, and the large errors do not have a significant influence on the performance. It indicates that our Taylor3DNet can reconstruct the 3D shape accurately without the need for higher-order terms. 

\begin{table}[t]
\scriptsize
\renewcommand{\arraystretch}{1}
\renewcommand{\tabcolsep}{1.2mm}
\centering
\begin{tabular}{c|ccccccc}
\toprule
Order                              & 0       & 1     & 2     & 3     & 4     & 5     & 6     \\ \hline
Error Mean($10^{-4}$)              & 109.79  & 6.50  & 1.42  & 0.90  & 0.66  & 0.50  & 0.48  \\
Large Error Rate(\textperthousand) & 421.955 & 1.999 & 0.027 & 0.012 & 0.011 & 0.011 & 0.011 \\
\bottomrule
\end{tabular}
\vspace{-2mm}
\caption{Experiments on representation capacity of Taylor series with different orders. We set an error threshold of $0.01$ to calculate the rate of the query points with an SDF error larger than it.} 
\label{tab:capacity}
\vspace{-4mm}
\end{table}

\subsection{Representation Capacity}
\label{sec:capacity}
In this section, we evaluate the upper limit capacity of our proposed representation. We randomly select 100 shapes from each category and sample 1000 landmark points for each shape. Then we sample a grid ($10^3$) of query points around each landmark point. All the shapes are normalized into a unit cube and the side length of the grid is 0.08. We apply the least square method to compute the coefficients of the Taylor series. Finally, we calculate the SDF of each query point by the obtained coefficients and analyze the SDF errors, which are shown in \cref{tab:capacity}.

We can observe that the error descending of an order higher than 2 is not significant. The large error rate shows that the rate of query points with SDF error larger than 0.01. At order 2, the rate of query points with SDF error larger than this threshold is $0.027$, which is small enough, and increasing the order does not lead to significant improvements. These experiments demonstrate that our Taylor-based implicit representation has a strong capacity to represent 3D shapes with the 2-order Taylor series. Besides, it also indicates that most local regions of the 3D shapes do not have large coefficients of higher-order terms.

\begin{table}[t]
\scriptsize
\renewcommand{\arraystretch}{1}
\renewcommand{\tabcolsep}{1.5mm}
\centering
\begin{tabular}{l|l|ccc|ccc}
\toprule
\multirow{2}{*}{Task} & \multirow{2}{*}{Method} & \multicolumn{3}{c|}{Resolution=128}        & \multicolumn{3}{c}{Resolution=256}                            \\ \cline{3-8} 
                   & & $T_{\text{eval}}$     & $T_{\text{mcube}}$ & $N_{\text{eval}}$ & $T_{\text{eval}}$ & $T_{\text{mcube}}$ & $N_{\text{eval}}$ \\ 
\hline
\multirow{2}{*}{SVR} 
& OccNet~\cite{mescheder2019occupancy}   & 0.412                 & 0.131             & 118K         & 2.100             & 1.105            & 395K   \\
& Taylor3DNet                &      0.069            &       0.121       &        21K        &      0.112        &      0.978       &    21K\\
\hline
\multirow{2}{*}{PCR} 
& ConvOccNet~\cite{peng2020convolutional}          & 0.343                 & 0.122             & 119K             & 1.742             & 1.014            & 397K   \\
& Taylor3DNet                & 0.028                 & 0.108             & 8K               & 0.041             & 0.912            & 8K   \\
\bottomrule
\end{tabular}
\caption{Shape reconstruction speed comparison on the single-view reconstruction (SVR) task and point cloud reconstruction (PCR) task. $T_{\text{eval}}$ and $T_{\text{mcube}}$ (seconds) denote the average time cost of points evaluation and marching cubes respectively, $N_{\text{eval}}$ denotes the average number of points evaluated by the network.}
\label{tab:time}
\end{table}

\subsection{Inference Speed}
To demonstrate the inference acceleration effect of our approach, we compare with two classical deep-implicit-function-based reconstruction methods, \ie, OccNet~\cite{mescheder2019occupancy} and ConvOccNet~\cite{peng2020convolutional}, on the single-view reconstruction (SVR) task and the point cloud reconstruction (PCR) task, respectively. We directly adopt their network architectures to implement our Taylor3DNet for each task, except replacing the final occupancy prediction layer with a Taylor coefficients prediction layer. By using the same architecture, we can better highlight the superiority of our representation in the inference phase. We conduct experiments on the whole ShapeNet test set and generate meshes at the two most commonly-used resolutions, \ie, 128 and 256, on the same Titan V GPU.

\begin{figure}[t]
  \centering
  \includegraphics[width=\linewidth]{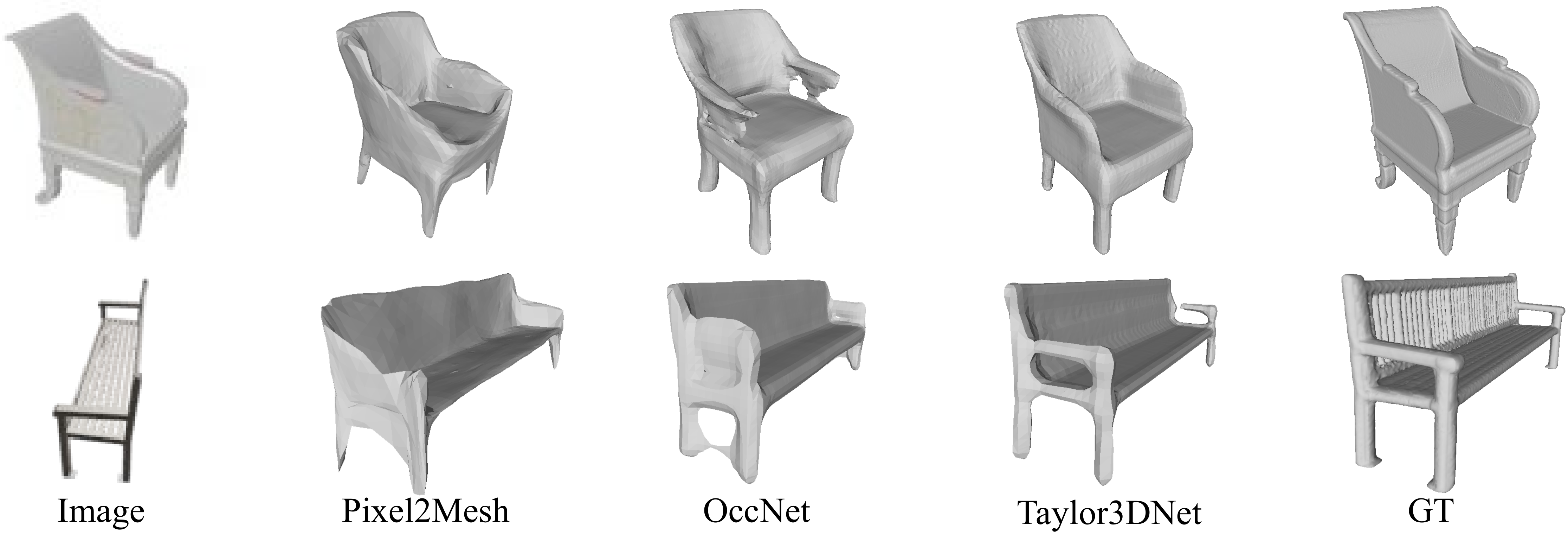}
  \caption{Qualitative comparison on single-view reconstruction.}
  \label{fig:svr_fig}
\end{figure}

\begin{figure*}[t]
  \centering
  \includegraphics[width=0.7\textwidth]{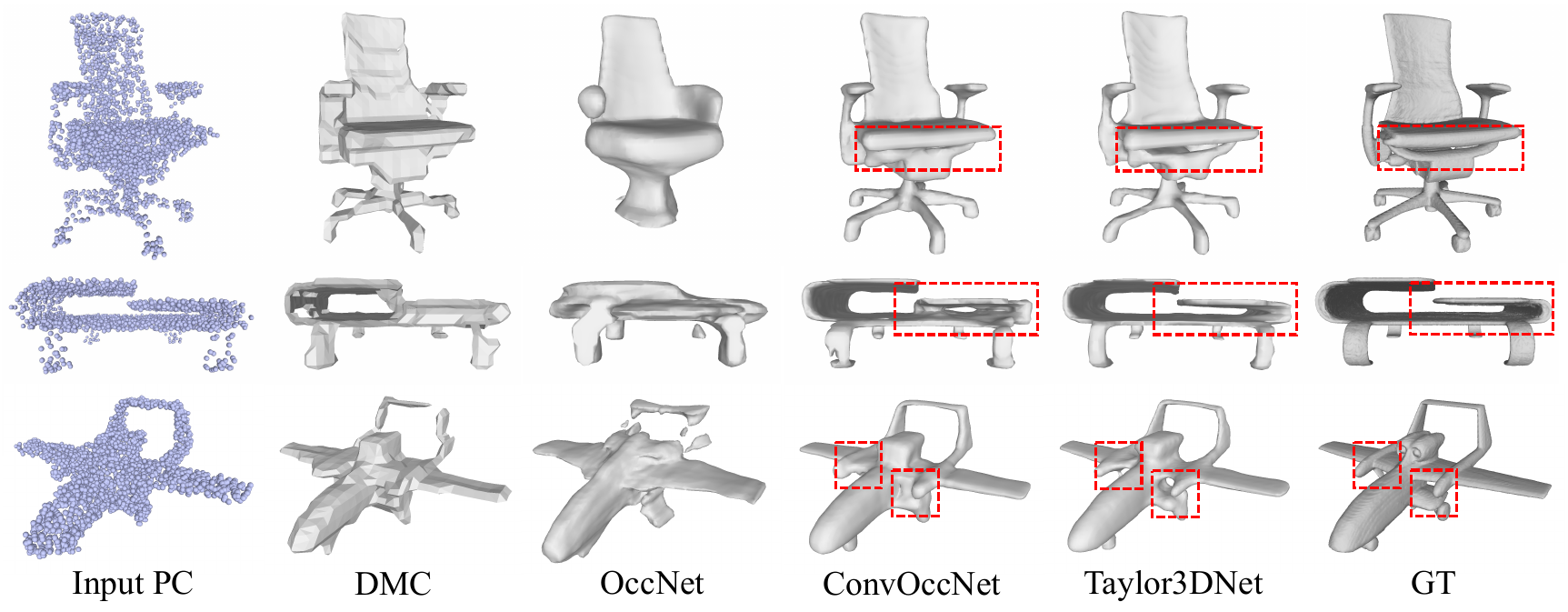}
  \caption{Qualitative comparison on the point cloud reconstruction task. We use red dotted boxes to highlight the significant differences compared with ConvOccNet.}
  \label{fig:pc_fig}
\end{figure*}

The complete shape reconstruction pipeline of deep implicit functions includes three steps: (i) feature encoding, (ii) points evaluation, and (iii) marching cubes. We can omit the feature encoding time since it is very short ($< 8$ms). In \cref{tab:time}, we present the average points evaluation time $T_{\text{eval}}$, average marching cubes time $T_{\text{mcube}}$ and the average number of points evaluated by the network $N_{\text{eval}}$. For our method, $T_{\text{eval}}$ is the sum of the network evaluation time for landmark points and the Taylor series computation time for all query points.

From \cref{tab:time}, we can observe that the points evaluation is the main bottleneck of mesh generation speed, and our method can significantly reduce the time of this step. By using the Multiresolution IsoSurface Extraction (MISE)~\cite{mescheder2019occupancy} algorithm, OccNet and ConvOccNet can reduce the number of points evaluated by the network from $128^3\approx 2$M to around 118K at the 128 resolution and from $256^3\approx 17$M to around $395$K at the 256 resolution, respectively. However, this is still no match for our reduction level. At both generation resolutions, we sample the landmark points from the resolution of 16 and refine them with our coarse-to-fine strategy, leading to only 21K landmark points for single-view reconstruction and 8K landmark points for point cloud reconstruction in total. 

Benefiting from the local modeling capability of the Taylor series, Taylor3DNet can exploit a fixed set of landmark points to represent a shape and evaluate any query point with naive Taylor series to extract arbitrary-resolution iso-surface. While showing similar reconstruction performance, Taylor3DNet reduces the points evaluation time by more than 90\% and the total reconstruction time by at least 65\% on both tasks.

Besides, our approach spends more time on single-view reconstruction than point cloud reconstruction, because identifying the regions near the surface is easier with point cloud input. 
we use looser $\epsilon_{in}$ and $\epsilon_{out}$ for the single-view input, leading to more sampled landmark points than that of point cloud input.

\begin{table}[t]
\scriptsize
\renewcommand{\arraystretch}{1.0}
\renewcommand{\tabcolsep}{0.7mm}
\centering
\begin{tabular}{l|ccccc}
\toprule
Metric & 3D-R2N2~\cite{choy20163d} & PSGN~\cite{fan2017point} & Pixel2Mesh~\cite{wang2018pixel2mesh} & OccNet~\cite{mescheder2019occupancy} & Ours \\ \hline
IoU $\uparrow$ & 0.500 & - & 0.480 & 0.593 & \textbf{0.599} \\
Chamfer-$L_1\downarrow$ & 0.246 & 0.215 & 0.216 & 0.194 & \textbf{0.192} \\
F-Score $\uparrow$ & 0.347 & 0.142 & 0.274 & 0.523 & \textbf{0.553} \\
\bottomrule
\end{tabular}
\caption{Quantitative comparison on single-view reconstruction.}
\label{tab:SVR_exp}
\end{table}

\subsection{Performance on Shape Reconstruction Tasks}
As we have demonstrated, Taylor3DNet speeds up the shape reconstruction process significantly. However, the reconstruction performance is also crucial and should not degrade to guarantee practical value in real applications. This section will show that Taylor3DNet achieves performance comparable to state-of-the-art methods on different reconstruction tasks.

\begin{table}[t]
\scriptsize
\renewcommand{\arraystretch}{1.0}
\renewcommand{\tabcolsep}{1mm}
\centering
\begin{tabular}{l|ccccc}
\toprule
Metric & PSGN \cite{fan2017point} & DMC \cite{liao2018deep} & OccNet \cite{mescheder2019occupancy} & ConvOccNet \cite{peng2020convolutional} & Ours \\
\hline
IoU $\uparrow$ & - & 0.733 & 0.772 & 0.870 & \textbf{0.874} \\
Chamfer-$L_1$ $\downarrow$ & 0.178 & 0.076 & 0.082 & 0.048 & \textbf{0.043} \\
F-Score $\uparrow$ & 0.180 & 0.790 & 0.799 & 0.933 & \textbf{0.944} \\
\bottomrule
\end{tabular}
\caption{Quantitative comparison on point clouds reconstruction.}
\label{tab:pc_exp}
\end{table}

\begin{table}[t]
\scriptsize
\renewcommand{\arraystretch}{1.0}
\renewcommand{\tabcolsep}{1mm}
\centering
\begin{tabular}{l|ccccc}
\toprule
\multicolumn{1}{l|}{Metric} & SAL~\cite{Atzmon_2020_CVPR} & SALD~\cite{atzmon2021sald} & IGR~\cite{gropp2020implicit} & Neural-Splines~\cite{williams2021neural} & Ours \\ 
\hline
Chamfer-$L_1$ $\downarrow$ & 2.359 & 0.414 & 0.\textbf{402} & 0.507 & 0.442 \\
NC $\uparrow$ & 0.787 & 0.918 & 0.920 & 0.901 & \textbf{0.922} \\
F-Score $\uparrow$ & 0.742 & 0.964 & 0.967 & \textbf{0.970} & 0.936 \\
\bottomrule
\end{tabular}
\caption{
Quantitative comparison with deep optimization-based surface reconstruction methods. 
We report the normal consistency (NC) instead of IoU following their practice.
SALD, IGR, and Neural-Splines also take the normals as input apart from points.}
\label{tab:optim}
\end{table}

\begin{table}[t]
\scriptsize
\renewcommand{\arraystretch}{1.0}
\renewcommand{\tabcolsep}{2.0mm}
\centering
\begin{tabular}{l|cccc}
\toprule
Metric & Input & OccNet \cite{mescheder2019occupancy} & ConvOccNet \cite{peng2020convolutional} & Ours \\
\hline
IoU $\uparrow$ & 0.631 & 0.703 & 0.752 & \textbf{0.761} \\
Chamfer-$L_1\downarrow$ & 0.136 & 0.110 & 0091 & \textbf{0.081} \\
F-Score $\uparrow$ & 0.440 & 0.656 & 0.729 & \textbf{0.755} \\
\bottomrule
\end{tabular}
\caption{Quantitative comparison on voxel super-resolution.}
\label{tab:voxel_exp}
\end{table}

\noindent\textbf{Single-view Reconstruction.}
We compare with 4 baseline methods, \ie, 3D-R2N2~\cite{choy20163d}, PSGN~\cite{fan2017point}, Pixel2Mesh~\cite{wang2018pixel2mesh} and OccNet~\cite{mescheder2019occupancy} on the single-view reconstruction task. The ShapeNet renderings provided by Choy et al.~\cite{choy20163d} are used as input.
We use the implementations in the OccNet codebase\footnote{\url{https://github.com/autonomousvision/occupancy\_networks}} for all baselines, and our model adopts the network architecture of OccNet directly. 

\cref{tab:SVR_exp} shows the quantitative results, and we can see that Taylor3DNet achieves similar performance with OccNet and outperforms other baselines. The results demonstrate that our proposed shape representation can effectively maintain the performance of deep implicit functions. Besides, the qualitative results in \cref{fig:svr_fig} also show that our method can represent the curved surface well.
Thanks to the higher-order terms of the Taylor series, our representation can model the high-order surface explicitly compared with the occupancy representation.

\noindent\textbf{Point Cloud Reconstruction.}
For the point cloud reconstruction task, we follow the setting of ConvOccNet~\cite{peng2020convolutional}. We randomly subsample 3000 points from the point cloud as input and apply Gaussian noise with zero mean and standard deviation of 0.005. 
We compare with PSGN~\cite{fan2017point}, DMC~\cite{liao2018deep}, OccNet~\cite{mescheder2019occupancy} and ConvOccNet~\cite{peng2020convolutional}, which are representative learning-based methods. Our Taylor3DNet adopts the same architecture as ConvOccNet \cite{peng2020convolutional}.

The quantitative results in \cref{tab:pc_exp} show that Taylor3DNet achieves similar performance with ConvOccNet and outperforms other baselines. Although we do not pursue performance improvements, Taylor3DNet indeed obtains slightly better metrics than ConvOccNet. Moreover, we can observe that Taylor3DNet and ConvOccNet can generate more plausible results than other baselines in \cref{fig:pc_fig}. The red boxes highlight the advantage of Taylor3DNet for generating flat and narrow shape parts or holes. 


Moreover, we also compare with deep optimization-based surface reconstruction methods~\cite{Atzmon_2020_CVPR,atzmon2021sald,gropp2020implicit,williams2021neural} considering their popularity and promising performance. Unlike learning-based methods trained on a training dataset and then generalized to test point clouds, deep optimization-based methods try to fit the shape directly from an unseen point cloud (maybe with normals if available). The optimization process often takes a long time for each shape, so we choose 200 models from 4 challenging ShapeNet categories, \ie, plane, car, chair, and table, to compare with them. The quantitative results are presented in \cref{tab:optim}, which demonstrate that our method is on par with them.

\noindent\textbf{Voxel Super-Resolution.}
This task aims to predict a high-resolution surface from a coarse voxel grid. We use the $32^3$ ShapeNet voxelizations provided by Choy et al.~\cite{choy20163d} as input. We compare with OccNet~\cite{mescheder2019occupancy} and ConvOccNet~\cite{peng2020convolutional}, and use the same network architecture as ConvOccNet~\cite{peng2020convolutional} too. \cref{tab:voxel_exp} shows the quantitative results. By replacing the occupancy-based representation with the Taylor-based representation, Taylor3DNet does not degrade the performance of ConvOccNet but slightly improves the metrics.

\begin{table}[t]
\scriptsize
\renewcommand{\arraystretch}{1.0}
\renewcommand{\tabcolsep}{2mm}
\centering
\begin{tabular}{l|ccccc}
\toprule
Metric & order=0 & order=1 & order=2 & order=3 & order=4 \\ \hline
IoU $\uparrow$ & 0.726 & 0.800 & 0.874 & \textbf{0.875} & \textbf{0.875} \\
Chamfer-$L_1\downarrow$ & 0.101 & 0.057 & 0.043 & 0.043 & \textbf{0.042} \\
F-Score $\uparrow$ & 0.762 & 0.890 & 0.944 & \textbf{0.945} & \textbf{0.945} \\
\bottomrule
\end{tabular}
\caption{Quantitative results of point cloud reconstruction using different orders of Taylor series. There is a gap between the 1-order and the 2-order representation.}
\label{tab:order}
\end{table}

\begin{figure}[t]
  \centering
  \includegraphics[width=\linewidth]{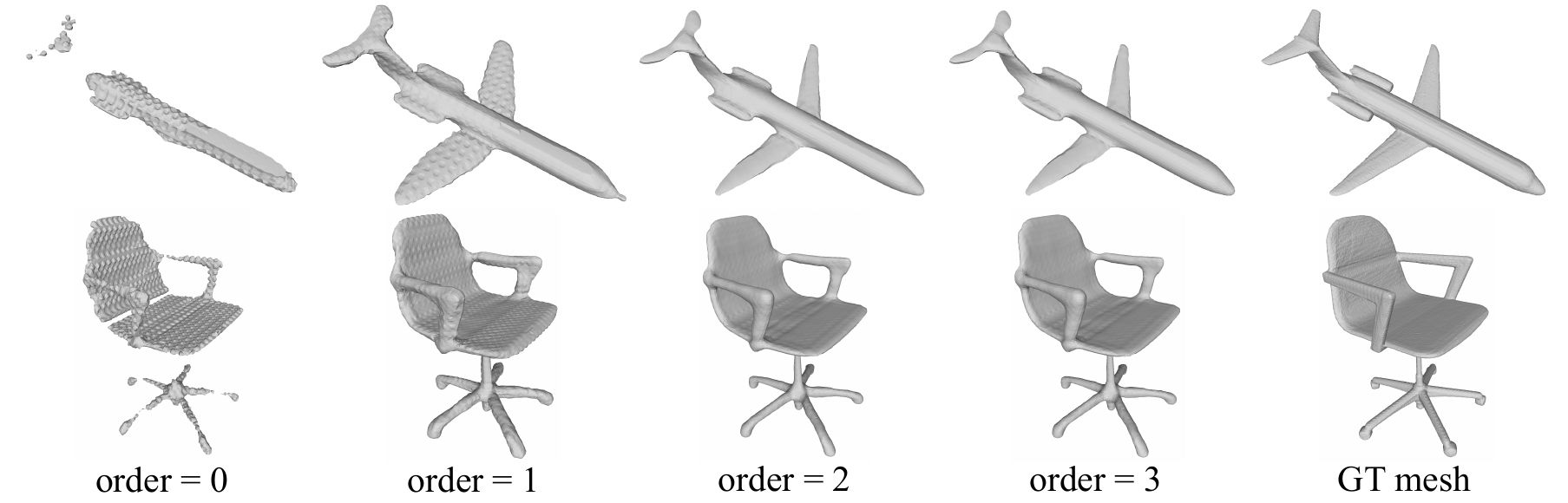}
  \caption{Qualitative results of point cloud reconstruction using different orders of Taylor series. The meshes generated with the 0-order or 1-order representations have artifacts, while those generated with the 2-order or 3-order representations look plausible. 
}
  \label{fig:order_fig}
\end{figure}

\begin{figure}[t]
  \centering
  \includegraphics[width=\linewidth]{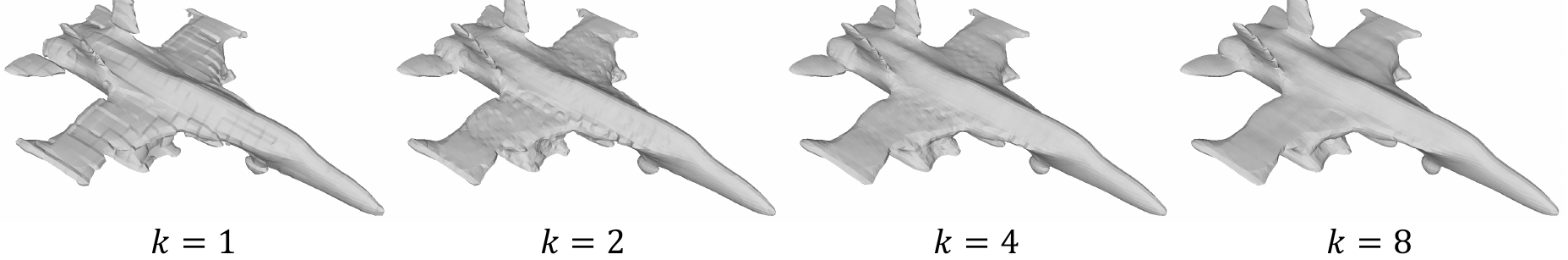}
  \caption{The visualization of the reconstruction results from a "plane" point cloud with varying numbers of expansion points at inference phase.
  }
  \label{fig:knn_fig}
\end{figure}

\begin{table}[!h]
\scriptsize
\renewcommand{\arraystretch}{1.0}
\renewcommand{\tabcolsep}{1.1mm}
\centering
\begin{tabular}{l|ccc|ccc}
\toprule
\multirow{2}{*}{k} & \multicolumn{3}{c|}{Single-view}                            & \multicolumn{3}{c}{Point  cloud}                            \\ \cline{2-7} 
                   & IoU$\uparrow$ &  Chamfer-$L_1\downarrow$ & F-Score$\uparrow$ & IoU$\uparrow$ &  Chamfer-$L_1\downarrow$ & F-Score$\uparrow$ \\ \hline
1                  & 0.593          & 0.193               & 0.543             & 0.867          & 0.045                  & 0.940             \\
2                  & 0.598          & \textbf{0.192}         & 0.551             & 0.872          & 0.044                  & 0.944             \\
4                  & 0.599 & \textbf{0.192}         & \textbf{0.553}    & 0.875 & \textbf{0.043}         & 0.945    \\
8                  & \textbf{0.600} & 0.196                  & 0.550             & \textbf{0.876} & \textbf{0.043}                  & \textbf{0.946}    \\    
\bottomrule
\end{tabular}
\caption{The influence of the number of nearest expansion points.}
\label{tab:KNN}
\vspace{-3mm}
\end{table}


\subsection{Ablation Studies}
In this section, we discuss two critical issues which affect the performance of Taylor3DNet. 

\noindent\textbf{Order of Taylor Series.}
After training a 4-order Taylor3DNet model on the point cloud reconstruction task, we generate meshes with different orders of the Taylor series. Specifically, when generating with the order $m$, we only compute the Taylor series with the terms of order $\leq m$ to evaluate the query points. From \cref{fig:order_fig} we can observe that the shapes reconstructed with the 0-order or 1-order Taylor series show severe artifacts and can only preserve a rough shape outline, while the shapes reconstructed with the 2-order Taylor series or higher look smooth and plausible. Besides, the quantitative results in \cref{tab:order} also show that order $<2$ leads to degraded performance while order $\ge 2$ achieves plausible performance. Since the differences between the orders greater than 2 are almost neglectable. We report our results with order 2 in all experiments.

\noindent\textbf{Number of Nearest Landmark Points.}
In the inference phase, we evaluate each query point by computing the weighted average of the Taylor series with $k$-nearest landmark points. We set $k=1, 2, 4, 8$ respectively and see how the performance is affected on both SVR and PCR tasks. \cref{tab:KNN} shows the quantitative results, from which we can see that using more landmark points and computing the weighted average among them promote the performance in general. However, when $k>4$, the performance degrades slightly on the SVR task because the landmark points far from the query point provide less accurate prediction and contribute negatively. Besides, exploiting larger $k$ leads to more memory usage. We empirically find $k=4$ is a cost-performance balanced choice and adopt it in all experiments. Moreover, we make a visualization of the meshes generated with different $k$ values in \cref{fig:knn_fig}. It shows that using a fair number of landmark points can significantly smooth the mesh surface and reduce the topology artifacts.

\section{Limitations and Future Work}\
Currently, the landmark points we leverage to represent shapes are statically sampled and cannot move, which wastes the representation ability of landmark points far from the surface. And we only conduct experiments on the synthetic shapes in ShapeNet, we think some extremely high-frequency details may be lost since we only use the low-order Taylor series to represent the shape. In future work, we hope to explore extending this approach to more complicated geometries and even non-watertight open surfaces, such as clothing.

\section{Conclusion}
In this work, we propose Taylor3DNet for fast 3D shape reconstruction, which utilizes a set of landmark points and their corresponding Taylor series' coefficients to represent a shape. The number of landmark points is independent of the resolution of iso-surface extraction. By predicting the coefficients of the Taylor series corresponding to the landmark points in the inference phase, we can simplify the evaluation of any query point as naive Taylor series computation with several nearest landmark points. The experimental results demonstrate that Taylor3DNet can significantly accelerate shape inference while maintaining the plausible performance of classical deep implicit functions.

{\small
\bibliographystyle{ieee_fullname}
\bibliography{egbib}
}


\clearpage
\clearpage
\newpage

\renewcommand\thesection{\Alph{section}}
\setcounter{section}{0}

\section{Implementation Details}
\label{sec:implementation}

\subsection{Network Architectures}

Taylor3DNet aims to accelerate the inference speed of deep implicit functions. To better demonstrate the acceleration effect, we directly adopt the network architectures of classical implicit-function-based methods for all reconstruction tasks, and we will illustrate the details below.

\begin{figure*}[t]
  \centering
  \includegraphics[width=0.45\linewidth]{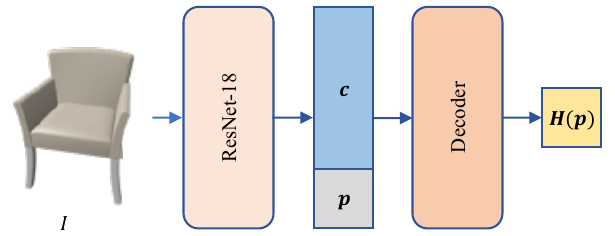}
  \caption{The overall architecture for single-view reconstruction. Given an image $I$ as input, a ResNet-18 encoder extracts a latent feature $\boldsymbol{c}$ from it. For each landmark point $\boldsymbol{p}$, $\boldsymbol{p}$ and $\boldsymbol{c}$ are fed into the decoder network to predict the coefficients of Taylor series $\boldsymbol{H}(\boldsymbol{p})$.}
  \label{fig:net_image}
\end{figure*}

\begin{figure}[t]
  \centering
  \includegraphics[width=\linewidth]{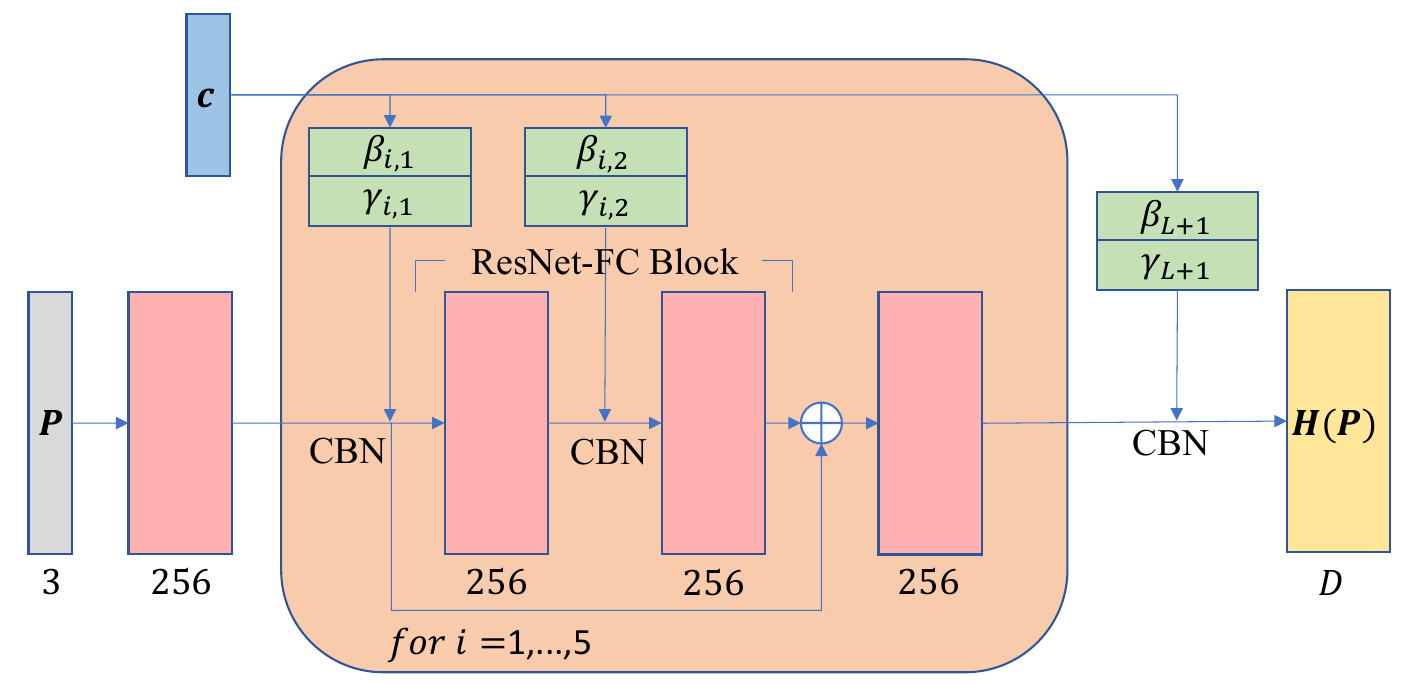}
  \caption{The decoder architecture for single-view reconstruction. Given a batch of landmark points $\boldsymbol{P}=\{\boldsymbol{p}_i\}_{i=1}^{N_{lk}}$ and the latent feature vector $\boldsymbol{c}$, the decoder outputs the coefficients of Taylor series corresponding to $\boldsymbol{P}$, \ie, $\boldsymbol{H}(\boldsymbol{p})=\{\boldsymbol{H}(\boldsymbol{p}_i)\}_{i=1}^{N_{lk}}$. $D$ is the dimension of Taylor series coefficients which is related to the order of Taylor series. \emph{CBN} denotes the Conditional Batch-Normalization~\cite{de2017modulating,dumoulin2016adversarially}, which is utilized in OccNet.} 
  \label{fig:decoder_onet}
\end{figure}

\noindent\textbf{Single-view Reconstruction.}
For the SVR task, we mainly compare with OccNet~\cite{mescheder2019occupancy}. We only replace the final occupancy prediction layer in the decoder with a Taylor series coefficients prediction layer. The overview of the network is shown in \cref{fig:net_image}, and the detailed decoder architecture is shown in \cref{fig:decoder_onet}.

\begin{figure*}[t]
  \centering
  \includegraphics[width=\linewidth]{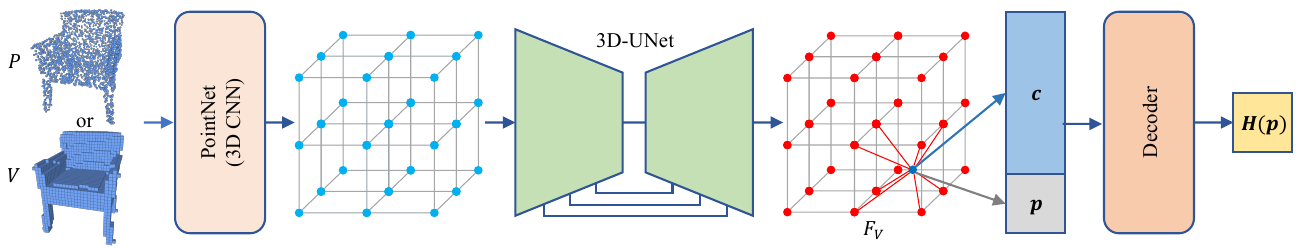}
  \caption{The overall architecture for point cloud reconstruction and voxel super-resolution. The network utilizes a shallow PointNet or a one-layer 3D CNN to obtain feature encoding from the point cloud input $P$ or voxelized input $V$. Then the feature encoding is fed into a 3D-UNet to construct a feature volume $F_V$. Given a landmark point $\boldsymbol{p}$, a latent feature $\boldsymbol{c}$ is extracted from $F_V$ by trilinear interpolation, then $\boldsymbol{p}$ and $\boldsymbol{c}$ are fed into the decoder to predict the coefficients of Taylor series $\boldsymbol{H}(\boldsymbol{p})$.}
  \label{fig:net_pc}
\end{figure*}

\begin{figure}[t]
  \centering
  \includegraphics[width=\linewidth]{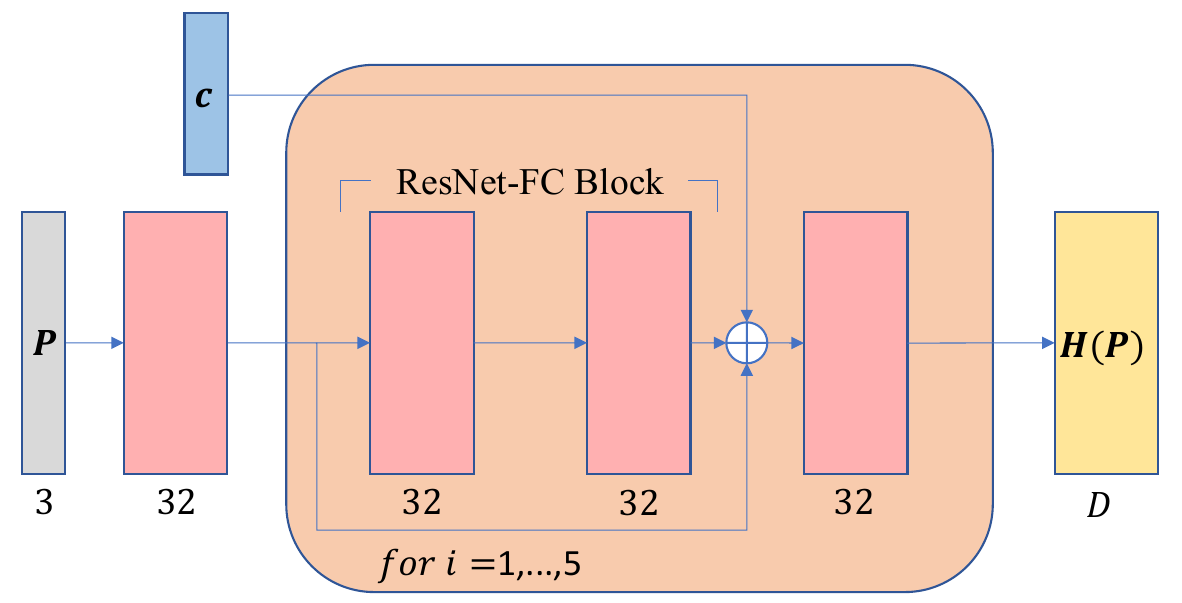}
  \caption{The decoder architecture for point cloud reconstruction and voxel super-resolution. Given a batch of landmark points $\boldsymbol{P}=\{\boldsymbol{p}_i\}_{i=1}^{N_{lk}}$ and the latent feature vectors $\boldsymbol{c}$ interpolated from the feature volume, the decoder outputs the coefficients of Taylor series corresponding to $\boldsymbol{P}$, \ie, $\boldsymbol{H}(\boldsymbol{P})=\{\boldsymbol{H}(\boldsymbol{p}_i)\}_{i=1}^{N_{lk}}$. $D$ is the dimension of Taylor series coefficients which is related to the order of Taylor series.}
  \label{fig:decoder_convonet}
\end{figure}

\noindent\textbf{Point Cloud Reconstruction.} For the task of surface reconstruction from point clouds, we mainly compare our approach with ConvOccNet \cite{peng2020convolutional}. Thus we adopt the same backbone as \cite{peng2020convolutional} and only modify the final output layer as well. The overview of the network is shown in \cref{fig:net_pc}, and the detailed decoder architecture is shown in \cref{fig:decoder_convonet}.


To be noted, ConvOccNet \cite{peng2020convolutional} has tested three types of encoders: (i) a single-plane encoder which projects the point features onto the ground plane, (ii) a multi-plane encoder which projects the point features onto three canonical planes, and (iii) a volume encoder which encodes the point features into a $32^3$ voxel grid. We use the volume encoder in our implementation because it can better represent the 3D information, and we empirically find that the plane encoders do not work well for our method. Besides, we also compare with ConvOccNet with a multi-plane encoder (denoted as CONet-2D) in \cref{sec:quantitative} and \cref{sec:qualitative}, since this architecture achieves the best quantitative results in the paper \cite{peng2020convolutional}.

\noindent\textbf{Voxel Super-Resolution.} Similar to point cloud reconstruction, we exploit the same ConvOccNet network \cite{peng2020convolutional} to tackle the task of voxel super-resolution. The only difference is that the shallow PointNet \cite{qi2017pointnet} at the beginning of the encoder is replaced with a one-layer 3D CNN to adapt to the voxelized inputs, as shown in \cref{fig:net_pc}.

\subsection{Training Details}
\label{sec:train}
We conduct our experiments on the ShapeNet~\cite{chang2015shapenet,choy20163d} subset containing 13 shape categories. We follow the preprocessing steps of OccNet~\cite{mescheder2019occupancy}\footnote{\url{https://github.com/autonomousvision/occupancy\_networks}} to obtain watertight meshes from the raw models. The watertight meshes are further normalized into a unit cube. For each mesh, we randomly sample 1024 points in the bounding volume and 3072 points near the surface as landmark points, and then evenly sample a $5^3$ grid of query points in a $0.08^3$ cube centered at each landmark point. We use the Kaolin~\cite{Kaolin}\footnote{\url{https://github.com/NVIDIAGameWorks/kaolin}} library to compute the ground truth SDF values of all query points for supervision.

We adopt the same train/val/test split as OccNet for all three reconstruction tasks. For single-view reconstruction and voxel super-resolution, we take the image renderings and $32^3$ voxelizations provided by Choy et.al~\cite{choy20163d} as input, respectively. Each input image is resized to $224\times 224$. And the point clouds provided by the OccNet~\cite{mescheder2019occupancy} training data are utilized for point cloud reconstruction. We set the batch size to 64 for single-view reconstruction and 32 for the other two tasks. For point cloud reconstruction and voxel super-resolution, we train the model with an initial learning rate of $1\times 10^{-3}$ for 180 epochs and divide the learning rate by 10 at the \nth{40} and \nth{120} epochs, respectively. For single-view reconstruction, we train with the same initial learning rate for 720 epochs and divide the learning rate by 10 at the \nth{360} and \nth{640} epochs, respectively.

\begin{figure*}[t]
  \centering
  \includegraphics[width=0.85\linewidth]{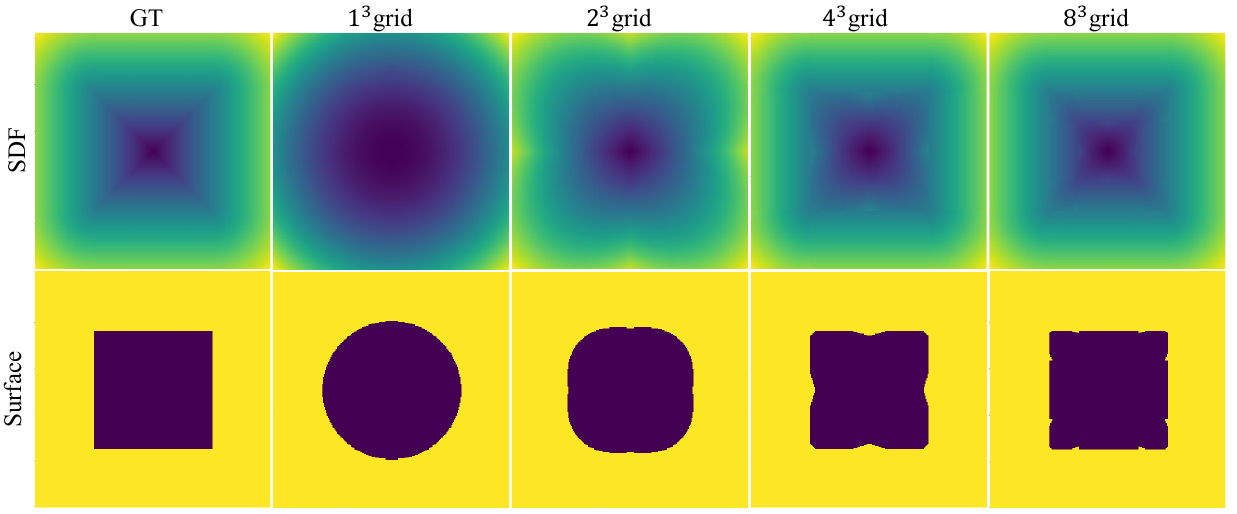}
  \vspace{-2mm}
  \caption{Representing a square with landmark points sampled at different resolutions. The upper row visualizes the obtained signed distance field, in which green areas are outside the surface and blue areas are inside the surface. The lower row visualizes the extracted iso-surface from the signed distance field.}
  \label{fig:sharp_corner}
\end{figure*}

\begin{figure*}[t]
  \centering
  \includegraphics[width=0.7\linewidth]{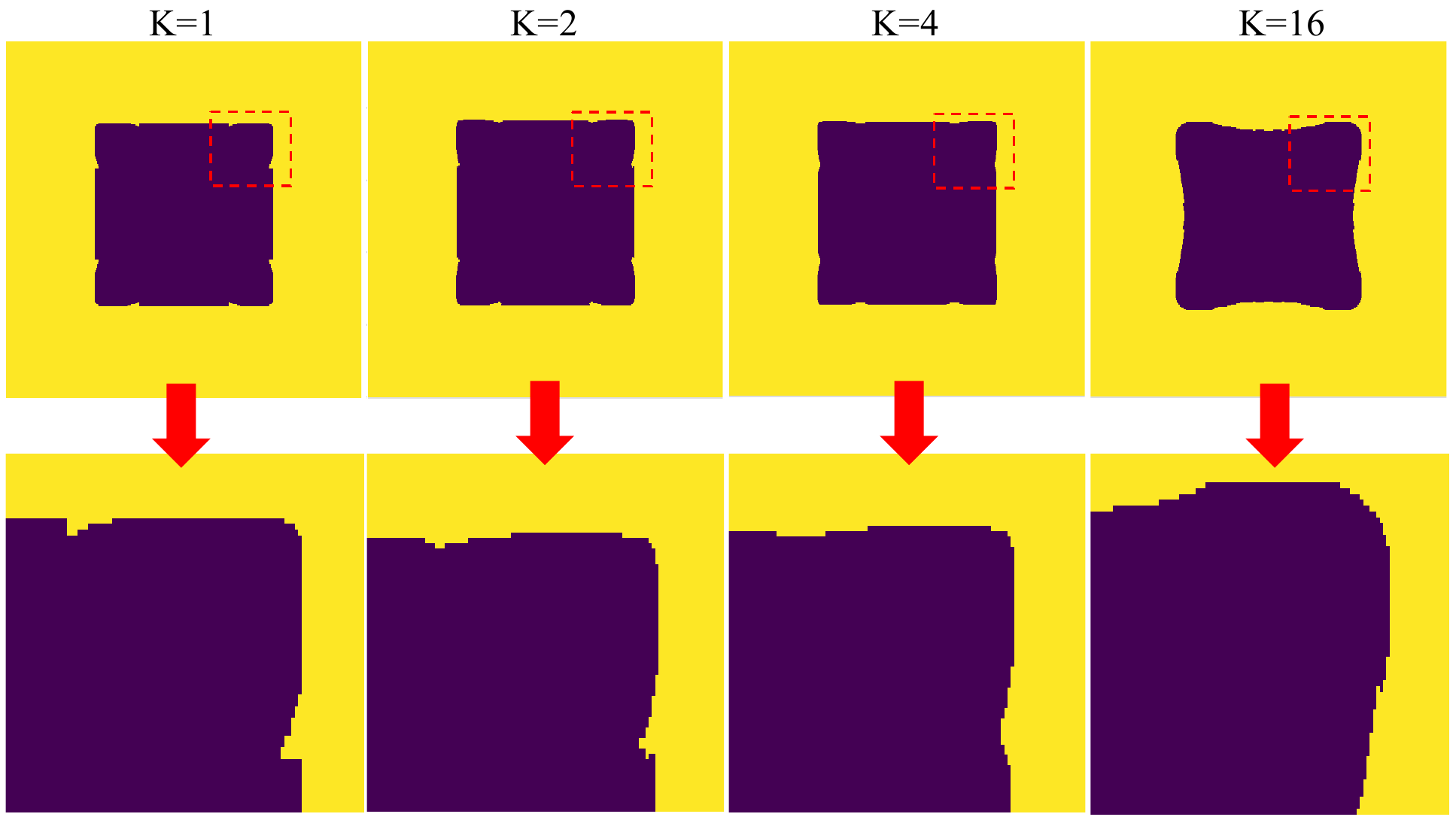}
  \vspace{-2mm}
  \caption{The square reconstruction results of using different numbers of nearest landmark points when evaluating the query points.}
  \label{fig:sharp_corner2}
  \vspace{-2mm}
\end{figure*}

\section{Additional Ablation Studies}
\label{sec:ablation}

\subsection{Representing Sharp Corners with Taylor Series}

Real-world objects often contain sharp corners. In this section, we use a toy 2D square shape to demonstrate the capacity of our Taylor-series-based representation for modeling sharp corners. Besides, we also show the effect of using different numbers of nearest landmark points to reconstruct the 2D shape.

\noindent\textbf{Sharp Corner Representation.} For the toy square shape, we sample a grid of landmark points at different resolutions to represent its signed distance field. We sample a set of query points near each landmark point and compute their ground truth SDF values. Then we regress the coefficients of the Taylor series of each landmark point with the Least-Squares algorithm. After that, we can compute the signed distance of any query point with the nearest landmark point and the Taylor series. We visualize the obtained signed distance field and the extracted iso-surface in the upper row and the lower row of \cref{fig:sharp_corner}, respectively.

We can observe that the denser landmark points we utilize, the better the sharp square shape can be represented. When sampling landmark points from the $1^3$ grid or the $2^3$ grid, each landmark point is demanded to represent a large area, thus the sharp corners cannot be reconstructed accurately. However, when sampling from the $4^3$ or $8^3$ grid, the signed distance field is very close to the ground truth, and the sharp corners can be approximated well.

\noindent\textbf{Number of Nearest Landmark Points.} Representing the square with an $8^3$ grid of landmark points, we also test how computing the signed distance of a query point with $K$ nearest landmark points affects the shape reconstruction results. The qualitative results are shown in \cref{fig:sharp_corner2}.

On the one side, using more nearest landmark points to evaluate a query point leads to a smoother predicted SDF field; on the other side, utilizing too many nearest landmark points will distort the reconstructed shape. Although using multiple nearest landmark points can smooth the extracted iso-surface, utilizing too many nearest landmark points will make the landmark points far from the query point participate in the SDF calculation and degrade the performance. Each landmark point can only represent a limited area, and the landmark points far from the query point provide inaccurate predictions. We adopt $K=4$ in our approach, which proves to be a practical and memory-efficient choice.

\begin{table}[t]
\footnotesize
\renewcommand{\arraystretch}{1}
\renewcommand{\tabcolsep}{1.8mm}
\centering
\begin{tabular}{l|ccc|c|c}
\toprule
\multirow{2}{*}{Metric} & \multicolumn{3}{c|}{order=1}  & order=2       & order=3 \\ \cline{2-6} 
                    & $D_l$=16  & $D_l$=32  & $D_l$=48  & $D_l$=16      & $D_l$=16    \\ \hline
IoU$\uparrow$       & 0.800     & 0.860     & 0.872     & 0.874         & 0.875     \\
Chamfer-$L_1\downarrow$ & 0.057 & 0.045     & 0.043     & 0.043         & 0.043     \\
F-Score$\uparrow$   & 0.890     & 0.942     & 0.944     & 0.944         & 0.945     \\
Memory (MB)         & 1486      & 1487      & 1534      & 1492          & 1616     \\
Eval Time (s)       & 0.032     & 0.071    & 0.274      & 0.032         & 0.035     \\
\bottomrule
\end{tabular}
\caption{The tradeoff between the density of landmark points and the order of the Taylor series. $D_l$ is the initial resolution for sampling landmark points in the coarse-to-fine sampling strategy.}
\label{tab:tradeoff}
\end{table}

\subsection{Density of Landmark Points \vs Order of Taylor Series}

In Section 4.5 of the manuscript, we have demonstrated that the reconstruction performance of using the 1-order Taylor series is much worse than using the 2-order or 3-order Taylor series. As discussed above, using denser landmark points can significantly improve the representation capacity. Intuitively, although the representation capacity of the 1-order Taylor series is limited, we can sample denser landmark points to improve the performance. 

We then experiment on the point cloud reconstruction task to see how many landmark points would the 1-order representation need to match the performance of the 2-order or 3-order representation. As \cref{tab:tradeoff} shows, the 1-order representation requires a denser landmark points grid ($>32^3$) to match the performance of the 2-order or 3-order representation ($16^3$), which inevitably increases the evaluation time. With a $16^3$ grid of landmark points, the 2-order or 3-order Taylor series leads to a marginal increment in memory usage while maintaining the inference speed. In conclusion, using the higher-order (2 or 3) Taylor series is superior to the 1-order Taylor series since they can represent a shape with much fewer landmark points. Besides, the performance difference between the 2-order and 3-order Taylor series can be neglected, thus we use order$=2$ in our experiments.

\begin{table*}[t]
\footnotesize
\renewcommand{\arraystretch}{1.15}
\renewcommand{\tabcolsep}{1.0mm}
\centering
\begin{tabular}{c|c|ccccccccccccc|c}
\toprule
\multirow{2}{*}{Metric} & \multirow{2}{*}{Method} & \multicolumn{13}{c|}{Category} & \multirow{2}{*}{Mean} \\ \cline{3-15} & & airplane & bench & cabinet & car & chair & display & lamp & speaker & rifle & sofa & table & phone & vessel  \\ 
\hline
\multirow{4}{*}{IoU $\uparrow$} 
& 3D-R2N2 \cite{choy20163d} & 0.427 & 0.369 & 0.672 & 0.660 & 0.452 & 0.462 & 0.304 & 0.625 & 0.370 & 0.629 & 0.435 & 0.625 & 0.490 & 0.500 \\
& Pix2Mesh \cite{wang2018pixel2mesh} &0.420 &0.323 &0.664 &0.552 &0.396 &0.490 &0.323 &0.599 &0.402 &0.613 &0.395 &0.661 &0.397 &0.480 \\
& OccNet \cite{mescheder2019occupancy} &0.591 &\bfseries 0.492 & \bfseries 0.750 &\bfseries 0.746 & \bfseries 0.530 &0.518 &0.400 &\bfseries 0.677 &0.480 &0.693 &0.542 &0.746 &0.547 &0.593 \\
& Ours & \bfseries 0.612 & 0.484 & 0.735 & 0.730 & 0.527 &\bfseries 0.543 & \bfseries 0.406 & 0.663 & \bfseries 0.533 & \bfseries 0.696 & \bfseries 0.555 &\bfseries 0.754 & \bfseries 0.567 &\bfseries 0.599 \\
\hline
\multirow{5}{*}{Chamfer-$L_1\downarrow$} 
& 3D-R2N2 \cite{choy20163d} & 0.201 & 0.206 & 0.217 & 0.214 & 0.266 & 0.301 & 0.504 & 0.316 & 0.185 & 0.223 & 0.241 & 0.187 & 0.229 & 0.246 \\
& PSGN \cite{fan2017point} &0.137 &0.181 &0.215 &0.169 &0.247 &0.284 &0.314 &0.316 &\bfseries 0.134 &0.224 &0.222 &0.161 &\bfseries 0.188 &0.215 \\
& Pix2Mesh \cite{wang2018pixel2mesh} &0.187 &0.201 &0.196 &0.180 &0.265 &0.239 &\bfseries 0.308 &0.285 &0.164 &0.212 &0.218 &0.149 &0.212 &0.216 \\
& OccNet \cite{mescheder2019occupancy} &0.134 &\bfseries 0.150 &0.153 &0.149 &\bfseries 0.206 &0.258 &0.368 &\bfseries 0.266 &0.143 &0.181 &0.182 &0.127 &0.201 &0.194 \\
& Ours & \bfseries 0.129 & 0.173 & \bfseries 0.152 & \bfseries 0.143 &0.227 &\bfseries 0.237 &0.425 &0.275 &0.163 &\bfseries 0.174 &\bfseries 0.180 &\bfseries 0.119 &0.226 &\bfseries 0.192 \\
\hline
\multirow{5}{*}{F-Score $\uparrow$} 
& 3D-R2N2 \cite{choy20163d} & 0.363 & 0.394 & 0.358 & 0.325 & 0.325 & 0.309 & 0.274 & 0.308 & 0.358 & 0.354 & 0.380 & 0.412 & 0.342 & 0.347 \\
& PSGN \cite{fan2017point} &0.301 &0.161 &0.071 &0.134 &0.077 &0.070 &0.120 &0.050 &0.391 &0.079 &0.106 &0.162 &0.193 &0.142 \\
& Pix2Mesh \cite{wang2018pixel2mesh} &0.487 &0.289 &0.182 &0.238 &0.226 &0.208 &0.268 &0.151 &0.462 &0.161 &0.315 &0.311 &0.261 &0.274 \\
& OccNet \cite{mescheder2019occupancy} & 0.624 &0.569 &0.595 &0.562 &0.426 &0.380 &0.388 &0.406 &0.575 &0.477 &0.582 &0.654 &0.419 &0.523 \\
& Ours & \bfseries 0.664 & \bfseries 0.598 &\bfseries 0.607 &\bfseries 0.593 &\bfseries 0.431 &\bfseries 0.422 &\bfseries 0.405 &\bfseries 0.422 &\bfseries 0.647 &\bfseries 0.502 &\bfseries 0.622 &\bfseries 0.694 &\bfseries 0.466 & \bfseries 0.553 \\
\bottomrule
\end{tabular}
\caption{Per-category quantitative results for single-view reconstruction on the ShapeNet~\cite{chang2015shapenet} dataset.}
\label{tab:image}
\end{table*}

\begin{table*}[t]
\footnotesize
\renewcommand{\arraystretch}{1.15}
\renewcommand{\tabcolsep}{1mm}
\centering
\begin{tabular}{c|c|ccccccccccccc|c}
\toprule
\multirow{2}{*}{Metric} & \multirow{2}{*}{Method} & \multicolumn{13}{c|}{Category} & \multirow{2}{*}{Mean} \\ \cline{3-15} & & airplane & bench & cabinet & car & chair & display & lamp & speaker & rifle & sofa & table & phone & vessel  \\ 
\hline
\multirow{5}{*}{IoU $\uparrow$} 
& DMC \cite{liao2018deep} & 0.653 & 0.605 & 0.856 & 0.779 & 0.739 & 0.813 & 0.601 & 0.856 & 0.645 & 0.855 & 0.701 & 0.880 & 0.712 & 0.733 \\
& OccNet \cite{mescheder2019occupancy} & 0.761 & 0.717 & 0.867 & 0.835 & 0.735 & 0.817 & 0.565 & 0.828 & 0.692 & 0.872 & 0.758 & 0.914 & 0.746 & 0.772 \\
& CONet-2D \cite{peng2020convolutional} & 0.848 & \bfseries 0.830 & \bfseries 0.940 & \bfseries 0.886 & \bfseries 0.871 & \bfseries 0.927 & 0.783 & 0.917 & 0.847 & \bfseries 0.936 & \bfseries 0.888 & \bfseries 0.953 & 0.865 & \bfseries 0.884 \\
& CONet-3D \cite{peng2020convolutional} & 0.847 & 0.790 & 0.922 & 0.877 & 0.853 & 0.902 & 0.790 & 0.913 & 0.827 & 0.923 & 0.860 & 0.941 & 0.859 & 0.870 \\
& Ours & \bfseries 0.850 & 0.788 & 0.922 & 0.881 & 0.861 & 0.908 & \bfseries 0.810 & \bfseries 0.918 & \bfseries 0.848 & 0.925 & 0.851 & 0.944 & \bfseries 0.871 & 0.874 \\
\hline
\multirow{6}{*}{Chamfer-$L_1\downarrow$} 
& PSGN \cite{fan2017point} & 0.127 & 0.156 & 0.195 & 0.154 & 0.210 & 0.197 & 0.242 & 0.237 & 0.115 & 0.182 & 0.196 & 0.140 & 0.160 & 0.178 \\
& DMC \cite{liao2018deep} & 0.069 & 0.071 & 0.072 & 0.101 & 0.072 & 0.063 & 0.91 & 0.081 & 0.062 & 0.063 & 0.070 & 0.044 & 0.081 & 0.076 \\
& OccNet \cite{mescheder2019occupancy} & 0.056 & 0.059 & 0.074 & 0.099 & 0.089 & 0.076 & 0.138 & 0.115 & 0.061 & 0.069 & 0.072 & 0.042 & 0.085 & 0.082 \\
& CONet-2D \cite{peng2020convolutional} & 0.034 & \bfseries 0.035 & \bfseries 0.046 & 0.075 & \bfseries 0.046 & \bfseries 0.037 & \bfseries 0.059 & 0.063 & 0.029 & \bfseries 0.041 & \bfseries 0.038 & \bfseries 0.027 & 0.043 & 0.044 \\
& CONet-3D \cite{peng2020convolutional} & \bfseries 0.033 & 0.041 & 0.054 & 0.080 & 0.049 & 0.042 & 0.068 & 0.065 & 0.031 & 0.046 & 0.043 & 0.030 & 0.045 & 0.048 \\
& Ours & \bfseries 0.033 & 0.041 & \bfseries 0.046 & \bfseries 0.057 & \bfseries 0.046 & 0.040 & \bfseries 0.059 & \bfseries 0.055 & \bfseries 0.028 & 0.042 & 0.043 & 0.028 & \bfseries 0.039 & \bfseries 0.043 \\
\hline
\multirow{6}{*}{F-Score $\uparrow$} 
& PSGN \cite{fan2017point} & 0.373 & 0.206 & 0.099 & 0.167 & 0.101 & 0.112 & 0.148 & 0.083 & 0.476 & 0.125 & 0.121 & 0.212 & 0.262 & 0.180 \\
& DMC \cite{liao2018deep} & 0.790 & 0.783 & 0.834 & 0.716 & 0.801 & 0.853 & 0.709 & 0.794 & 0.831 & 0.850 & 0.810 & 0.944 & 0.748 & 0.790 \\
& OccNet \cite{mescheder2019occupancy} & 0.865 & 0.862 & 0.856 & 0.758 & 0.753 & 0.813 & 0.614 & 0.738 & 0.837 & 0.845 & 0.840 & 0.942 & 0.757 & 0.799 \\
& CONet-2D \cite{peng2020convolutional} & 0.965 & \bfseries 0.965 & \bfseries 0.956 & 0.849 & 0.939 & \bfseries 0.971 & 0.891 & 0.892 & 0.980 & \bfseries 0.953 & \bfseries 0.968 & 0.988 & 0.930 & 0.942 \\
& CONet-3D \cite{peng2020convolutional} & \bfseries 0.968 & 0.944 & 0.931 & 0.833 & 0.930 & 0.956 & 0.910 & 0.880 & 0.969 & 0.942 & 0.952 & 0.987 & 0.927 & 0.933 \\
& Ours & 0.964 & 0.948 & 0.943 & \bfseries 0.876 & \bfseries 0.942 & 0.962 & \bfseries 0.921 & \bfseries 0.902 & \bfseries 0.981 & \bfseries 0.953 & 0.955 & \bfseries 0.991 & \bfseries 0.943 & \bfseries 0.944 \\
\bottomrule
\end{tabular}
\caption{Per-category quantitative results for point cloud reconstruction on the ShapeNet~\cite{chang2015shapenet} dataset. CONet-2D denotes ConvOccNet~\cite{peng2020convolutional} with a 2D multi-plane encoder, while CONet-3D denotes ConvOccNet with a 3D volume encoder.}
\label{tab:pc}
\end{table*}

\begin{table*}[t]
\footnotesize
\renewcommand{\arraystretch}{1.15}
\renewcommand{\tabcolsep}{1mm}
\centering
\begin{tabular}{c|c|ccccccccccccc|c}
\toprule
\multirow{2}{*}{Metric} & \multirow{2}{*}{Method} & \multicolumn{13}{c|}{Category} & \multirow{2}{*}{Mean} \\ \cline{3-15} & & airplane & bench & cabinet & car & chair & display & lamp & speaker & rifle & sofa & table & phone & vessel  \\ 
\hline
\multirow{5}{*}{IoU $\uparrow$} 
& Input & 0.498 & 0.504 & 0.772 & 0.695 & 0.622 & 0.685 & 0.497 & 0.788 & 0.472 & 0.759 & 0.582 & 0.722 & 0.607 & 0.631 \\
& OccNet \cite{mescheder2019occupancy} & 0.721 & 0.610 & 0.811 & 0.807 & 0.664 & 0.714 & 0.512 & 0.786 & 0.612 & 0.802 & 0.617 & 0.806 & 0.679 & 0.703 \\
& CONet-2D \cite{peng2020convolutional} & 0.746 & \bfseries 0.655 & \bfseries 0.845 & 0.814 & 0.731 & 0.778 & 0.631 & 0.831 & 0.673 & 0.836 & 0.679 & 0.828 & 0.731 & 0.752 \\
& CONet-3D \cite{peng2020convolutional} & 0.745 & 0.640 & 0.841 & \bfseries 0.816 & 0.744 & 0.768 & 0.651 & 0.834 & 0.670 & 0.835 & 0.678 & 0.821 & 0.736 & 0.752 \\
& Ours & \bfseries 0.766 & 0.652 & 0.841 & 0.815 & \bfseries 0.751 & \bfseries 0.782 & \bfseries 0.655 & \bfseries 0.835 & \bfseries 0.689 & \bfseries 0.838 & \bfseries 0.680 & \bfseries 0.836 & \bfseries 0.751 & \bfseries 0.761 \\
\hline
\multirow{5}{*}{Chamfer-$L_1\downarrow$} 
& Input & 0.127 & 0.122 & 0.147 & 0.188 & 0.130 & 0.129 & 0.139 & 0.152 & 0.123 & 0.129 & 0.130 & 0.120 & 0.135 & 0.136 \\
& OccNet \cite{mescheder2019occupancy} & 0.069 & 0.084 & 0.112 & 0.109 & 0.115 & 0.118 & 0.187 & 0.150 & 0.080 & 0.101 & 0.111 & 0.080 & 0.114 & 0.110 \\
& CONet-2D \cite{peng2020convolutional} & 0.056 & 0.070 & \bfseries 0.096 & 0.110 & 0.087 & 0.086 & 0.111 & 0.118 & 0.063 & 0.084 & 0.086 & 0.069 & 0.087 & 0.092 \\
& CONet-3D \cite{peng2020convolutional} & 0.055 & 0.074 & 0.099 & 0.110 & 0.083 & 0.091 & \bfseries 0.098 & 0.118 & 0.064 & 0.086 & 0.086 & 0.072 & 0.086 & 0.091 \\
& Ours & \bfseries 0.051 & \bfseries 0.068 & 0.097 & \bfseries 0.098 & \bfseries 0.079 & \bfseries 0.083 & 0.099 & \bfseries 0.116 & \bfseries 0.057 & \bfseries 0.082 & \bfseries 0.084 & \bfseries 0.064 & \bfseries 0.079 & \bfseries 0.081 \\
\hline
\multirow{5}{*}{F-Score $\uparrow$} 
& Input & 0.411 & 0.470 & 0.435 & 0.338 & 0.452 & 0.461 & 0.439 & 0.455 & 0.409 & 0.470 & 0.468 & 0.486 & 0.420 & 0.440 \\
& OccNet \cite{mescheder2019occupancy} & 0.806 & 0.727 & 0.638 & 0.713 & 0.606 & 0.585 & 0.511 & 0.545 & 0.742 & 0.651 & 0.636 & 0.744 & 0.622 & 0.656 \\
& CONet-2D \cite{peng2020convolutional} & 0.855 & 0.779 & 0.689 & 0.717 & 0.709 & 0.694 & 0.684 & 0.625 & 0.814 & 0.717 & 0.710 & 0.788 & 0.705 & 0.735 \\
& CONet-3D \cite{peng2020convolutional} & 0.859 & 0.749 & 0.668 & 0.716 & 0.726 & 0.662 & 0.716 & 0.618 & 0.812 & 0.703 & 0.699 & 0.764 & 0.707 & 0.729 \\
& Ours & \bfseries 0.882 & \bfseries 0.792 & \bfseries 0.691 & \bfseries 0.745 & \bfseries 0.752 & \bfseries 0.712 & \bfseries 0.732 & \bfseries 0.635 & \bfseries 0.855 & \bfseries 0.724 & \bfseries 0.717 & \bfseries 0.829 & \bfseries 0.750 & \bfseries 0.755 \\
\bottomrule
\end{tabular}
\caption{Per-category quantitative results for voxel super-resolution on the ShapeNet~\cite{chang2015shapenet} dataset. CONet-2D denotes ConvOccNet~\cite{peng2020convolutional} with a 2D multi-plane encoder, while CONet-3D denotes ConvOccNet with a 3D volume encoder.}
\label{tab:voxel}
\end{table*}

\begin{figure*}[t]
  \centering
  \includegraphics[width=0.7\linewidth]{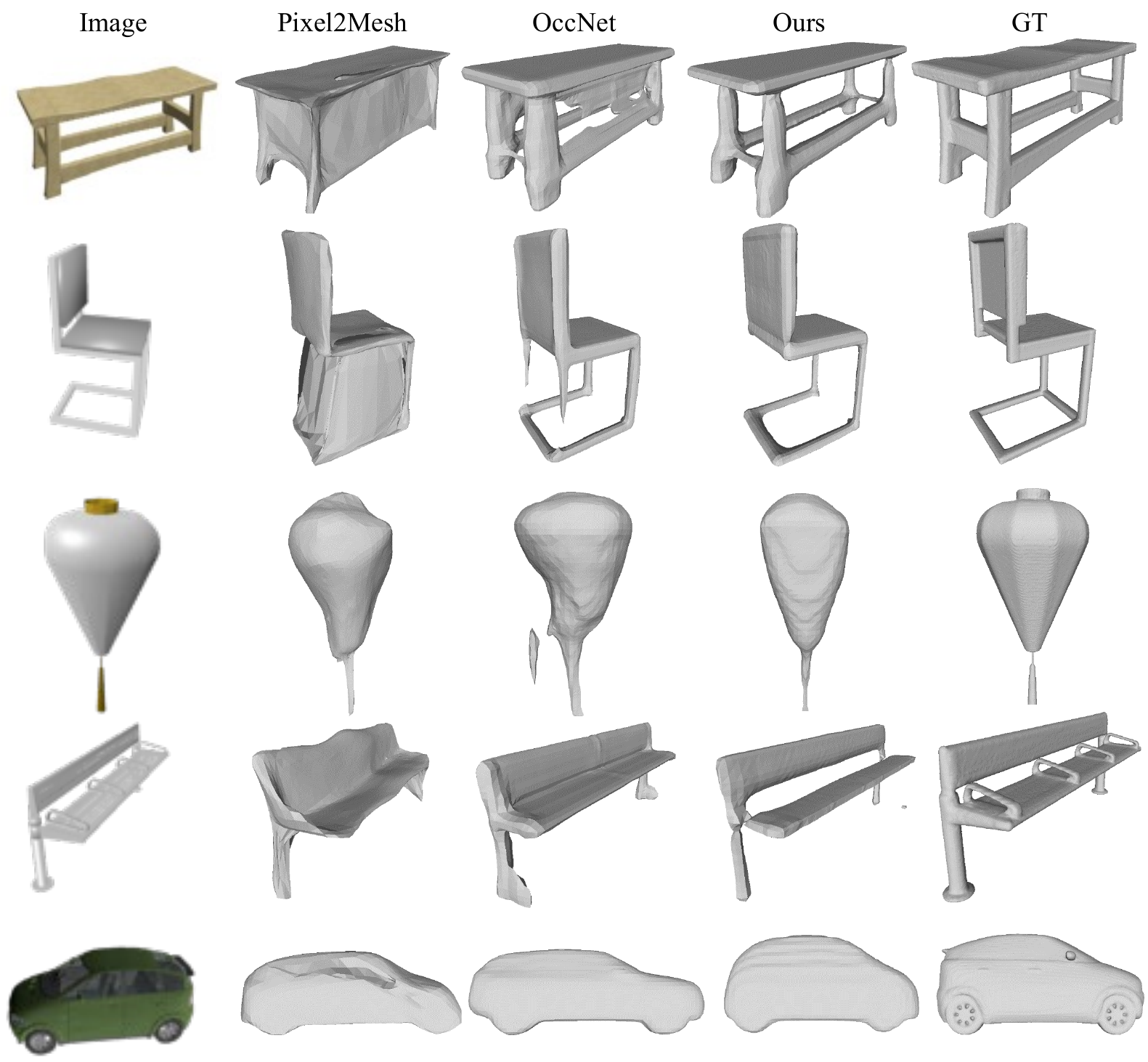}
  \caption{Additional qualitative results for single-view reconstruction.}
  \label{fig:image_results}
\end{figure*}

\section{Detailed Quantitative Results}
\label{sec:quantitative}
\cref{tab:image}, \cref{tab:pc} and \cref{tab:voxel} show the detailed quantitative results of the three different reconstruction tasks, respectively. From the numerics, we can observe that our method generally performs on par with or slightly better than the baseline methods. Although the CONet-2D baseline, \ie, ConvOccNet~\cite{peng2020convolutional} with a multi-plane encoder obtains higher IoU than our approach on the point cloud reconstruction task, we still outperform it in terms of Chamfer-$L_1$ distance and F-Score. On the task of voxel super-resolution, our method achieves considerable performance gains with the same backbone network as CONet-3D.

\section{Additional Qualitative Results}
\label{sec:qualitative}

\begin{figure*}[t]
  \centering
  \includegraphics[width=0.95\linewidth]{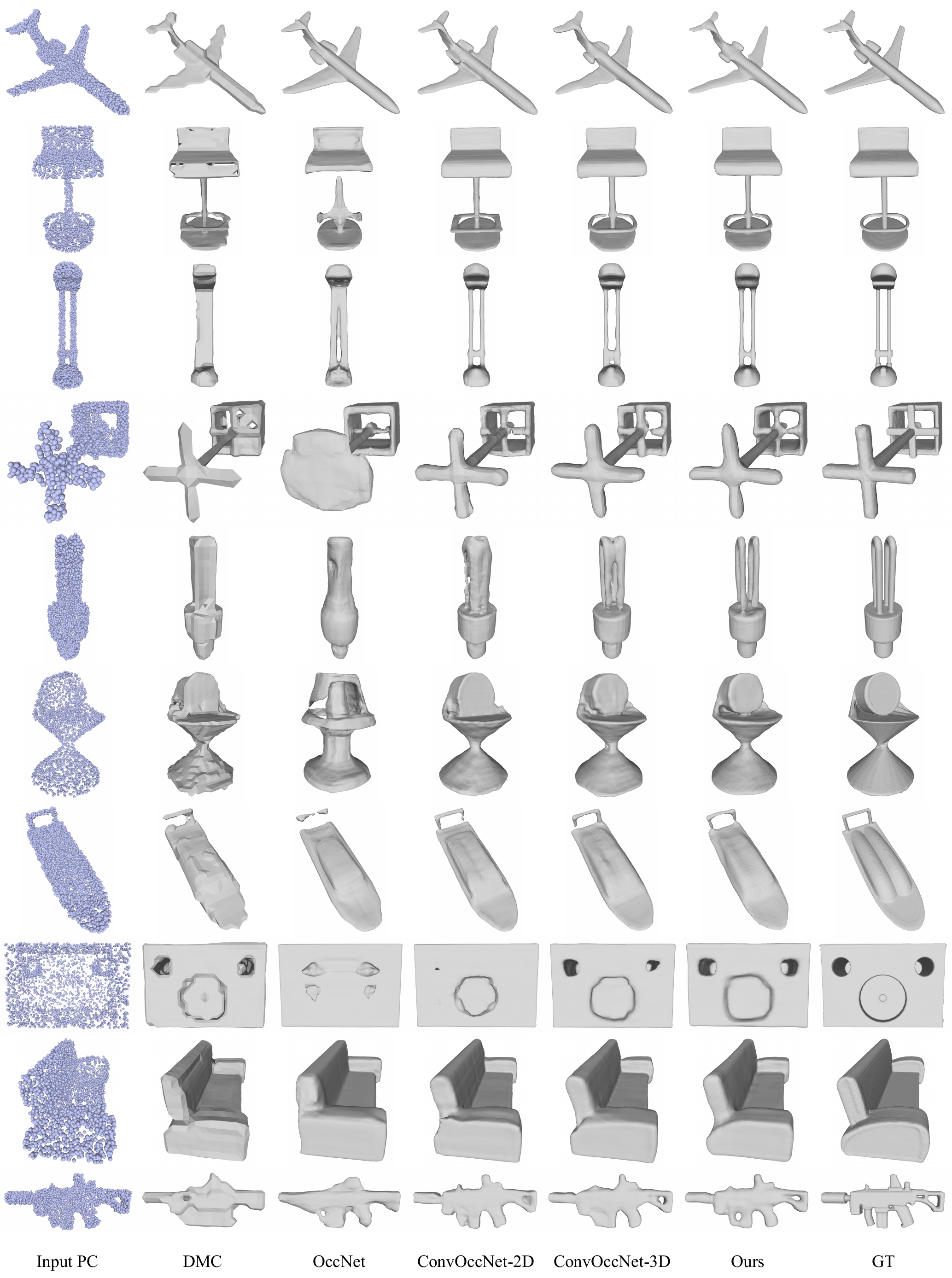}
  \caption{Additional qualitative results for point cloud reconstruction.}
  \label{fig:pc_results}
\end{figure*}

\begin{figure*}[t]
  \centering
  \includegraphics[width=0.95\linewidth]{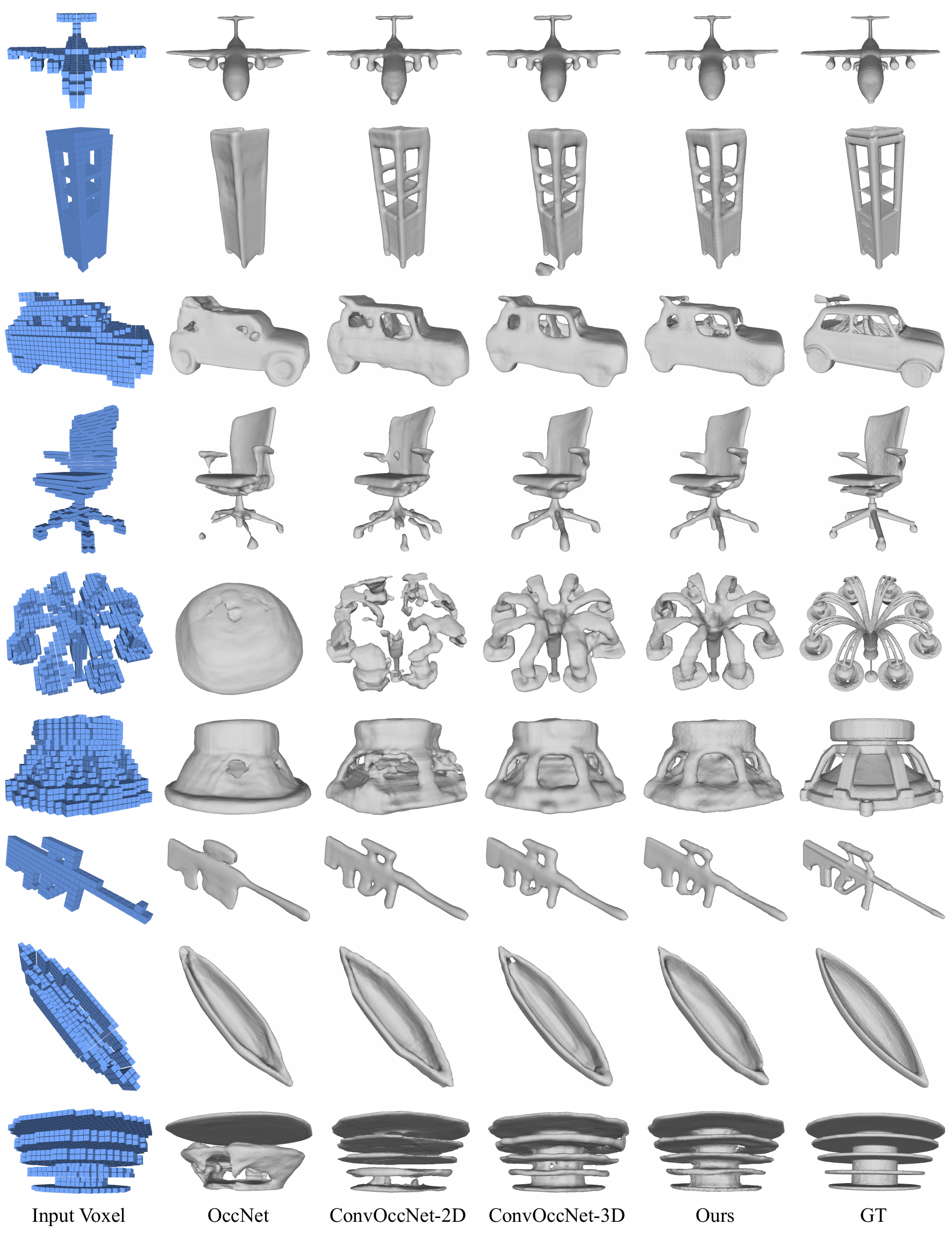}
  \caption{Additional qualitative results for voxel super-resolution.}
  \label{fig:voxel_results}
\end{figure*}

We also provide extensive qualitative reconstruction results in this section. Specifically, the single-view reconstruction results are presented in \cref{fig:image_results}, and the result from point clouds and voxels are presented in \cref{fig:pc_results} and \cref{fig:voxel_results}, respectively. We can observe that our method can reconstruct smooth and complete surfaces and preserve complicated geometric details from the visualizations. Even better, our method can reconstruct high-resolution shapes at a much higher inference speed than the baseline methods, as shown in the manuscript. The results validate the representation capacity of our method.

\end{document}